\documentclass[a4paper,fleqn]{cas-dc}

\usepackage[numbers]{natbib}
\def\tsc#1{\csdef{#1}{\textsc{\lowercase{#1}}\xspace}}
\tsc{WGM}
\tsc{QE}

\usepackage{amssymb}

\makeatletter
\usepackage{amsmath}
\usepackage{multirow}
\usepackage{graphicx}
\usepackage[ruled,linesnumbered]{algorithm2e}
\usepackage{array}

\usepackage[dvipsnames]{xcolor}
\usepackage{stmaryrd} % for brackets like [[ ]]
\usepackage[justification=centering]{caption} % caption居中
\usepackage{subcaption}

\newcolumntype{C}[1]{>{\centering\arraybackslash}p{#1 em}}

\newdefinition{definition}{Definition}
\newtheorem{theorem}{Theorem}
\newtheorem{corollary}{Corollary}
\newproof{proof}{Proof}
\renewcommand{\cite}[1]{\citep{#1}}

\AtBeginDocument {%
          \def\resetMathstrut@{%
           \setbox\z@\hbox{\the\textfont\symoperators\char40}%
           \ht\Mathstrutbox@\ht\z@ \dp\Mathstrutbox@\dp\z@}%
}%
\makeatother

\DeclareMathOperator*{\E}{\mathbb{E}}
\newcommand*{\dif}{\mathop{}\!\mathrm{d}}

\newcommand{\nosection}[1]{\vspace{1.5pt}\noindent\textbf{#1.}}

\begin{document}
\title[mode = title]{Towards Secure and Practical Machine Learning via Secret Sharing and Random Permutation}
\author[zju]{Fei Zheng}
\author[zju]{Chaochao Chen}
\author[zju]{Xiaolin Zheng}
\cormark[1]

\author[jzt]{Mingjie Zhu}

\affiliation[zju]{
    organization={College of Computer Science and Technology, Zhejiang  University, Hangzhou, China}
}
\affiliation[jzt]{
     organization={JZTData Technology, Hangzhou, China}
 }

\cortext[1]{Corresponding author.}

\shorttitle{Towards Secure and Practical Machine Learning via Secret Sharing and Random Permutation}
\shortauthors{Fei Zheng, Chaochao Chen, Xiaolin Zheng, Mingjie Zhu}

\begin{abstract}
% Background
With the increasing demand for privacy protection, privacy-preserving machine learning has been drawing much attention from both academia and industry. 
% Motivation
However, most existing methods have their limitations in practical applications. 
% Examples
On the one hand, although most cryptographic methods are provable secure, they bring heavy computation and communication. 
On the other hand, the security of many relatively efficient privacy-preserving techniques (e.g., federated learning and split learning) is being questioned, since they are non-provable secure. 
% Our method
Inspired by previous works on privacy-preserving machine learning, 
we build a privacy-preserving machine learning framework by combining random permutation and arithmetic secret sharing via our compute-after-permutation technique.  
% Efficiency 
Our method is more efficient than existing cryptographic methods, since it can reduce the cost of element-wise function computation. 
% Security
Moreover, by adopting distance correlation as a metric for evaluating privacy leakage, we demonstrate that our method is more secure than previous non-provable secure methods. 
Overall, our proposal achieves a good balance between security and efficiency. 
% Expriment results
Experimental results show that our method not only is up to $5\times$ faster and reduces up to 80\% network traffic compared with state-of-the-art cryptographic methods, but also leaks less privacy during the training process compared with non-provable secure methods. 
\end{abstract}

\maketitle

\section{Introduction}\label{section:intro}
%% Topic introduce
Machine learning has been widely used in many real-life scenarios in recent years. 
In many practical cases, a good machine learning model requires data from multiple sources (parties). 
% Example
For example, two hospitals want to use their patients' data to train a better disease-diagnosis model, and two banks want to use their clients' data to train a more intelligent credit-ranking model.
However, in many cases, the data sources are unwilling to share their data since their data is valuable or contains user privacy.
Hence, how to train a model while keeping the privacy of sensitive data becomes a major challenge.

%% Background
%
In traditional machine learning scenarios, data is \textit{centralized} in a server or a cluster for model training. 
Two settings are usually encountered when privacy is taken into account~\cite{yang2019federated}.
One is that different data sources have different samples with the same set of features. In other words, data is \textit{horizontally} distributed. 
Another setting is that data is \textit{vertically} distributed, i.e., different data sources have overlapped samples but different features.
Federated learning~\cite{mcmahan17fed} mainly focuses on the horizontal setting. 
Privacy-Preserving Machine Learning (PPML) systems, e.g., ~\cite{mohassel2017secureml, wagh2019securenn, mohassel2018aby3}, work for both settings via multiple data sources sharing data to one or several servers. 
In this paper, we aim to build a PPML system under the 3-server setting, which is suitable for both vertically and horizontally distributed data and supports both model training and inference. 
%
% An overview of our system is in Figure \ref{fig:my_system}.

\begin{figure}[ht]
    \centering
    \includegraphics[width=6cm]{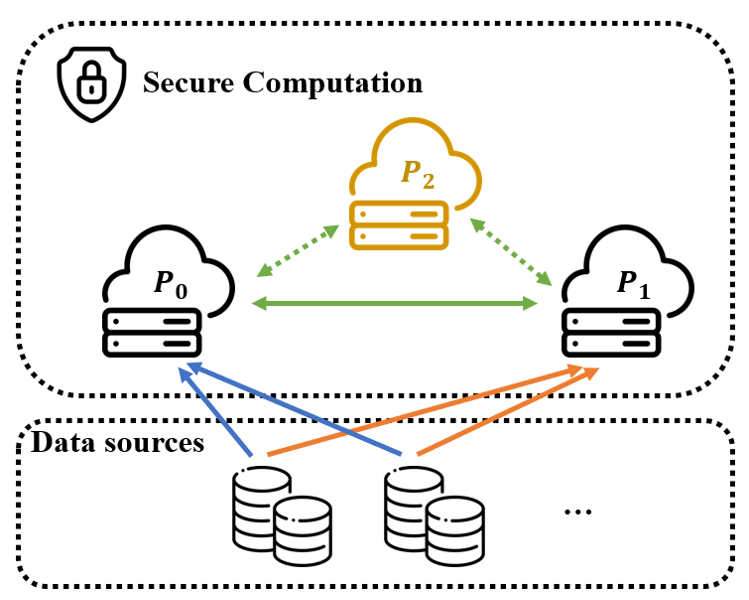}

    \caption{System overview.}
    \label{fig:my_system}
\end{figure}

%% Existing work summary
In recent years, various methods have been proposed for privacy-preserving machine learning.
These methods can be generally classified into two classes: \textit{provable secure} methods and \textit{non-provable secure} methods. 
Using provable secure methods, the adversary cannot derive any information about the input data within polynomial time under given threat models, 
e.g., \textit{semi-honest} (passive) threat model and \textit{malicious} (active) threat model~\cite{evans2017pragmatic}.
In contrast, non-provable secure methods would leak certain information about the input data.
Provable secure methods  mainly use cryptographic primitives including Homomorphic Encryption (HE)~\cite{paillier1999, gentry2009fhe}, Garbled Circuits (GC)~\cite{yao1986gc}, Secret Sharing (SS)~\cite{shamir1979share}, 
or other customized secure Multi-Party Computation (MPC)  protocols~\cite{mohassel2017secureml, demmler2015aby,wagh2019securenn, mohassel2018aby3,chen2021homomorphic,fang2021large} to realize basic operations (e.g., addition, multiplication, comparison) of machine learning. Hence, their security can be formally proved. 
Non-provable secure methods include federated learning~\cite{mcmahan17fed}, split learning~\cite{vepa2018split}, and some hybrid methods~\cite{zhangqiao2019gelunet,xie2019bayhenn,he2020transnet,zheng2020ppdnn,zheng2021asfgnn}. Those methods are usually simple in theory and easy to implement.
% Differential privacy is out of the scope
Another widely used technique is differential privacy~\cite{dwork2014dp}, which protects privacy by adding certain noise in the specific stage of training or inference. 
However, it has a trade-off between privacy and accuracy and is mainly adopted in centralized or horizontal scenarios, and thus is out of the scope of this paper.

%% Shortcomings of existing works
Although privacy-preserving machine learning has been widely studied, most existing methods have their limitations in practical applications.
Cryptographic methods require heavy computation and communication and are always difficult to implement. 
Non-provable secure methods are usually very efficient, but they may leak privacy under certain conditions. Also, they lack quantification for privacy leakage. 
Existing researches indicate that some methods may leak information that can be exploited by the adversary~\cite{abuadbba2020splitcnn, harry2020exposing}.

%% Our solution
In this paper, we propose a practical method based on secret sharing and random permutation, which is more efficient than existing cryptographic methods and more secure than most non-provable secure methods.
% Overview
We present the system overview in Figure \ref{fig:my_system}. 
From it, we can see that our method adopts a 3-server setting, i.e., the secure computation in our method involves three servers: $P_0$, $P_1$, and $P_2$. 
First, all data sources upload their data to $P_0$ and $P_1$ in a secret-shared manner, which means that neither $P_0$ or $P_1$ has any meaningful information of the original data but random values. 
Then all three servers interactively perform model training or inference in an encrypted (secret-shared) manner. 
Finally, the desired output is reconstructed in plaintext.

Technically, our proposal adopts Arithmetic Secret Sharing (A-SS) scheme from previous work on MPC~\cite{demmler2015aby} and improves efficiency via the compute-after-permutation technique. 
% Basic idea
For linear operations like addition and multiplication, our method behaves like~\cite{wagh2019securenn, riazi2018chameleon} where $P_2$ is used to generate beaver triples for secure multiplication. 
For non-linear element-wise functions like the widely used Sigmoid and ReLU, our method lets $P_2$ perform the computation via compute-after-permutation technique. 
Briefly speaking, $P_2$ gets a random permutation of the input from $P_0$ and $P_1$, then computes the result of the element-wise function. After that, $P_2$ sends the shares of the computation result back to $P_0$ and $P_1$.

% Advantage
% Efficiency
The compute-after-permutation technique exploits the element-wise property of many activation functions. 
Unlike MPC methods, our method distributes the computation for element-wise functions to a server ($P_2$), hence greatly reduces computation and communication costs.
% Privacy
To guarantee security, instead of sending plaintext to $P_2$, a random permutation of the input is sent. 
we claim that the adversary can hardly extract any information of the original data from the compute-after-permutation technique, mainly due to two reasons. 
First, the number of permutations grows exponentially as the number of elements grows.
Second, the input is already a random transformation of the original data (input of the model) since it have passed at least one neural network layer.
Moreover, by quantitative analysis based on the statistical metric distance correlation~\cite{szekely2007dc} and simulated experiments, we show that our method leaks even less information than compressing the data into only one dimension. 
Compared with non-provable secure methods like split learning~\cite{vepa2018split} which directly reveal hidden representations, our method leaks far less privacy in terms of distance correlation.

% Result
The experiment results show that our method has less computation time and network traffic in logistic regression and neural network models compared with state-of-the-art cryptographic methods.
Moreover, our method achieves the same accuracy as centralized training.

%% Contribution
We summarize our main contributions in this paper as follows.

\begin{itemize}
    \item We propose a secure and practical method for privacy-preserving machine learning, based on arithmetic sharing and the compute-after-permutation technique.
    \item We quantify privacy leakage and demonstrate the security of our method by using statistic measure distance correlation. We show that our method is more secure than existing non-provable secure methods.
    \item We benchmark our method with existing PPML systems on different models, and the results demonstrate that our method has less network communication and running time than centralized plaintext training, while achieving the same accuracy.
\end{itemize}

\section{Related Work}
We divide the methods for privacy-preserving machine learning into two groups. 
One is cryptographic methods that use cryptographic primitives to build PPML systems. 
Another is non-provable secure methods whose security cannot be proved in a cryptographic sense and may leak intermediate results during model training.

\subsection{Cryptographic Methods}
Cryptographic methods are based on cryptographic primitives such as GC, SS, HE, and MPC protocols. 
The security of these methods can be formally proved under their settings, i.e., no adversary can derive any information of the original data within polynomial time under the specific security setting.

The privacy problem is encountered with at least two parties.
Hence, many PPML systems are built on a 2-server setting, where two servers jointly perform the computations of model inference or training.
For example, CryptoNets~\cite{bachrach2016cryptonets} first used Fully Homomorphic Encryption (FHE) for neural network inference.
ABY~\cite{demmler2015aby} provided a two-party computation protocol that supports computations including addition, subtraction, multiplication, and boolean circuit evaluation, by mixing arithmetic, boolean, and Yao sharing together. 
Motivated by ABY, SecureML~\cite{mohassel2017secureml} and MiniONN~\cite{liu2017minionn} mixed A-SS and GC together to implement privacy-preserving neural networks.
Gazelle~\cite{juvekar2018gazelle} avoided expensive FHE by using packed additive HE to improve efficiency and used GC to calculate non-linear activation functions.

The above two-party protocols are usually not efficient enough for practical applications, and many PPML systems used the 3-server or 4-server setting recently, where three or four parties jointly perform the PPML task.
% Property of 3PC
Under the 3-server setting, GC and HE are sometimes avoided due to their inefficiency. 
%
% And to perform non-linear computation, customized protocols are used.
%
% 3PC methods
SecureNN~\cite{wagh2019securenn} used a party to assist the most significant bit and multiplication on arithmetic shared values. 
ABY3~\cite{mohassel2018aby3} performed three types of sharing in \cite{demmler2015aby} under 3 parties to improve efficiency. 
Chameleon~\cite{riazi2018chameleon} used a semi-honest third-party for beaver triple generation and oblivious transfer.
\cite{ajith2019astra, ajith2020blaze, ajith2020flash, ajith2020trident, ajith2020swift}
developed new protocols based on arithmetic and boolean sharing under 3-server or 4-server settings and claim to outperform previous methods. 
Open-sourced PPML libraries, such as CryptFlow~\cite{kumar2020cryptflow}, TF-Encrypted~\cite{chenghong2020tfe}, and Crypten~\cite{knott2020crypten} are also based on the semi-honest 3-servers setting where a certain party is usually used to generate Beaver triples.

% Conclusion of cryptographic methods
Overall, cryptographic methods are aimed to design protocols that achieve provable security. 
They usually use SS as the basis, and develop new protocols for non-linear functions to improve efficiency.
%
% The comparison function is the most important, since it can used to interpolate other non-linear functions.

\subsection{Non-provable Secure Methods}
We summarize the methods whose security cannot be formally proved in a cryptographic sense into non-provable secure methods. 
Those methods may use non-cryptographic algorithms and usually reveal certain intermediate results.

Federate learning and split learning are two well-known non-cryptographic methods.
% Federated
Federated learning~\cite{mcmahan17fed} is commonly used in scenarios where data is horizontally distributed among multiple parties. 
It protects data privacy via sending model updates instead of original data to a server. 
% Split 
Split learning~\cite{gupta2018distributed, vepa2018split} is a straightforward solution for vertically distributed data.
It simply splits the computation graph into several parts, while a server usually maintains the middle part of the graph as a coordinator.
% Other methods
There are also other methods, for example:
\cite{he2020transnet} designed a transformation layer that uses linear transformations with noises added on the raw data to protect privacy;
\cite{zhangqiao2019gelunet} used additive HE for matrix multiplication but outsourced the activation function to different clients while keeping them unable to reconstruct the original input;
\cite{xie2019bayhenn} used a bahyesian network to generate noisy intermediate results to protect privacy based on the \textit{learning with errors} problem~\cite{regev2009lwe}.

Non-provable methods usually leak certain intermediate results from joint computation.
For example, federated learning leaks the model gradients, 
and split learning leaks the hidden representations.
Their security depends on the security of certain intermediate results.
However, those intermediate results can be unsafe.
For example, \cite{zhuligeng2019deepleakage, yinhongxu2021gradients} show that the gradients can be used to reconstruct the original training samples,
and \cite{harry2020exposing} proved methods of \cite{zhangqiao2019gelunet, xie2019bayhenn} can leak model parameters in certain cases.
There have also been several attempts on the security of split learning.
In this paper, since we focus on the computation of hidden layers, we compare our proposal with split learning in Section \ref{section:security}. 
\section{Preliminaries}
In this section, we review the fundamental techniques of our method, including arithmetic secret sharing, random permutation, and distance correlation. 

\subsection{Arithmetic Secret Sharing}
% Secret sharing introduction
The idea of Secret Sharing (SS)~\cite{shamir1979share} is to distribute a secret to multiple parties, and 
the secret can be reconstructed only when more than a certain number of parties are together. 
Hence, SS is widely used in MPC systems where 2 or 3 parties are involved.
% SS -> A-SS
Arithmetic Secret Sharing (A-SS) is a kind of secret sharing that behaves in an arithmetic manner.
When using A-SS, it is simple to perform arithmetic operations such as addition and multiplication on shared values.
Due to its simplicity and information-theoretic security under semi-honest settings, it is widely used in different kinds of PPML systems~\cite{riazi2018chameleon, wagh2019securenn, mohassel2018aby3}.

% Detail of our arithmetic sharing
We use the A-SS scheme proposed by \cite{demmler2015aby} where a value is additively shared between two parties. 
The addition of two shared values can be done locally, while the multiplication of two shared values requires multiplicative triples introduced by Beaver~\cite{beaver1991}. 
To maintain perfect security, arithmetic sharing is performed on the integer ring $\mathbb Z_{2^L}$. 
The arithmetic operations we refer below are also defined on that ring by default.

%% Basic operations
% The basic operations are described as follows:

\textbf{Share: }
A value $x$ is shared between two parties $P_0, P_1$, which means that $P_0$ holds a value $\langle x\rangle_0$ while $P_1$ holds a value $\langle x\rangle_1$ such that $\langle x\rangle_0 +\langle x\rangle_1 = x$.

To share a value $x\in \mathbb Z_{2^L}$, party $P_i$ (can be $P_0, P_1, P_2$ or any other party) simply picks $r \stackrel{\$}\leftarrow \mathbb Z_{2^L}$ and sends $r$ and $x - r$ to $P_0$ and $P_1$ respectively. 

\textbf{Reconstruct: }
$P_0$ and $P_1$ both send their shares $\langle x\rangle_0, \langle x\rangle_1$ to some party $P_i$. 
The raw value $x = \langle x\rangle_0 + \langle x\rangle_1$ is reconstructed immediately on $P_i$ by summing up the two shares.

\textbf{Add: }
When adding a \textit{public value} (values known to all parties) $a$ to a shared value $x$, $P_0$ maintains $\langle x\rangle_0 + a$ and $P_1$ maintains $\langle x\rangle_1$.
When adding a shared value $a$ to a shared value $x$, two party add their shares respectively,
\textit{i.e.}, $P_0$ gets $\langle x\rangle_0 + \langle a\rangle_0$ and $P_1$ gets $\langle x\rangle_1 + \langle a\rangle_1$.

\textbf{Mul/MatMul: }
% shared x public
When multiplying a shared value $x$ with a public value $a$, the two parties multiply their shares of $x$ with $a$ respectively, 
i.e., $P_0$ gets $\langle x\rangle_0 \cdot a$, $P_1$ gets $\langle x\rangle_1 \cdot a$.
% shared x shared
For multiplication of shared values, we use $P_2$ to generate beaver triples. 
When multiplying shared values $x$ and $y$, $P_2$ generates beaver triples $u$, $v$, $w$,
where $u$ and $v$ are randomly picked from $\mathbb Z_{2^L}$ such that $w = u\cdot v$, and then shares them to $P_0$ and $P_1$.
Then $P_0$ and $P_1$ reconstruct $x - u$ and $y - v$ to each other. 
After that, $P_0$ calculates $(x-u)(y-v) + \langle(x-u)v\rangle_0 + \langle u(y-v)\rangle_0 + \langle w\rangle_0$, $P_1$ calculates $\langle(x-u)v\rangle_1 + \langle u(y-v)\rangle_1 + \langle w\rangle_1$. 
Then they get their shares of the result $xy$. 
% Extending to MatMul
Multiplication of matrices (MatMul) is performed similarly.

\subsection{Random Permutation}
% Introduction
Random permutation is used for privacy-preserving in many fields like 
data analysis~\cite{du2001ppsa}, linear programming~\cite{dreier2011ppmlp}, clustering~\cite{vaidya2003ppkmeans}, support vector machine~\cite{maekawa2018ppsvm, maekawa2019ppsvm} and neural network~\cite{he2020transnet}.
%
% Effectiveness
It can be efficiently executed in a time complexity of $\mathcal O(n)$~\cite{durstenfeld1964permutation}.

% Security
The security of random permutation is based on that the number of possibilities grows exponentially with the number of elements.
% Simple Example
For example, for a vector of length 20, there are $2.43\times10^{18}$ possible permutations, which means that the chances for the adversary to guess the original vector is negligible. 
% More detailed
We will present a more detailed analysis on the security of our proposed permutation method in Section \ref{section:security}.

\subsection{Distance Correlation}
% Introduction
Distance correlation~\cite{szekely2007dc} is used in statistics to measure the dependency between to two random vectors.
% Definition
For two random vectors $X\in\mathbb R^p$ and $Y \in \mathbb R^q$, their distance correlation is defined as follows:
\begin{equation}
    \text{Dcor}(X, Y) = \int_{\mathbb R^{p+q}} |f_{X,Y}(x, y) - f_X(x)f_Y(y)|^2 g(x,y)dxdy,
\end{equation}
where $f$ is the charateristic function and $g$ is a certain non-negative weight function. 
The distance correlation becomes 0 when two random vectors are totally independent and 1 when one random vector is an orthogonal projection of the other one.

% Estimation
Let $X$ and $Y$ be $n$ samples drawn from two distributions, their distance correlation can be estimated by 
\begin{equation}
    \text{Dcor}(X, Y)) \approx \dfrac{V_n^2(X,Y)}{\sqrt{V_n^2(X,X) V_n^2(Y, Y)}},
\end{equation}
where $V_n^2(X, Y) = \dfrac{1}{n^2}\sum_{k,l=1}^n A_{kl}B_{kl}$ is the estiamted distance covariance between $X, Y$ (similar for $V_n^2(X, X)$ and $V_n^2(X, Y)$).
$A_{kl}$ is the doubly-centered distances between $A_k$ and $A_l$, and so is $B_{kl}$.

% Why use distance correlation
We use distance correlation as the measure of privacy leakage because it is concrete in the theory of statistics, easy to estimate, 
and intuitively reflects the similarity of topological structure between two datasets. 
% Random projection related
Also, it is closely related to the famous Jonhson-Lindenstrauss lemma~\cite{johnson1984lemma}, 
which states that random projections approximately preserve the distances between samples hence reserve the utility of the original data~\cite{bingham2001rp}. 
\section{The Proposed Method}

In this section, we first overview the high-level architecture of our method.
Then we describe the building blocks of our method, i.e., fixed-point arithmetic and the compute-after-permutation technique.
Finally, we combine the building blocks to realize the inference and training of neural networks.

\subsection{Overview}
\begin{figure}[t]
\centering
    \begin{subfigure}[b]{0.43\textwidth}
    \includegraphics[width=8cm]{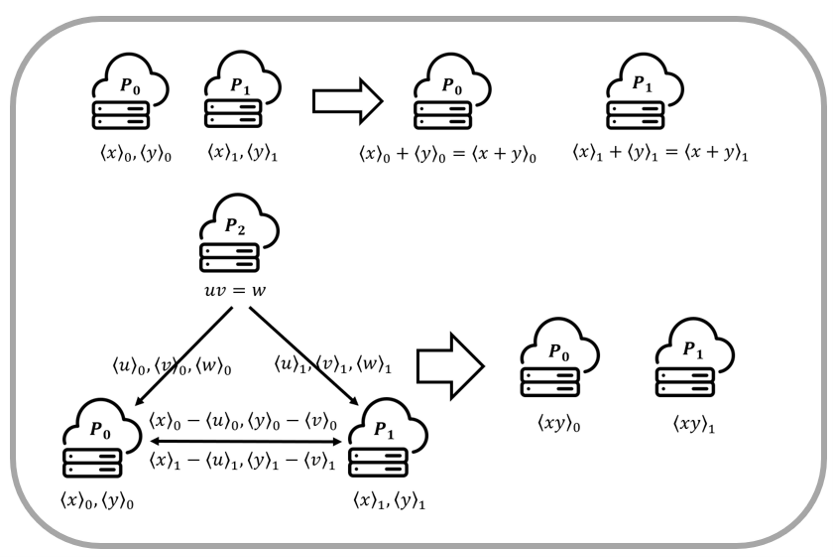}
    \subcaption{A-SS part: addition \& multiplication}
    \end{subfigure}

    \begin{subfigure}[b]{0.43\textwidth}
    \includegraphics[width=8cm]{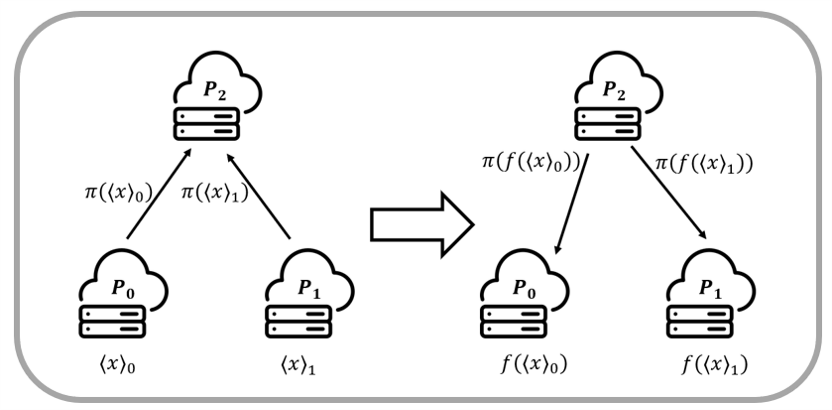}
    \subcaption{Non-linear part: non-linear element-wise functions}
    \end{subfigure}

\caption{Technique overview of our method.}
\label{fig:overview}
\end{figure}

% Describe
Our method combines A-SS and random permutation to maintain the balance between privacy and efficiency.
Similar to existing work \cite{wagh2019securenn, mohassel2018aby3}, our method is based on the 3-server setting, where data is shared between two servers $P_0$ and $P_1$, and is computed with the assist of $P_2$, as has been illustrated in Section \ref{section:intro}. 
Figure \ref{fig:overview} displays the high-level architecture of our method.

Since we use A-SS throughout the computation, all the values have to be converted to fixed-point.
This is done via dividing the 64-bit integer into higher 41 bits and lower 23 bits.
When converting a float-point value to fixed-point, the higher 41 bits are used for the integer part and the lower 23 bits are used for the decimal part. 
Linear operations like Add and Sub can be performed in an ordinary way, while the multiplication of shared values needs to be specially treated.
We develop the ShareClip algorithm to avoid the error introduced by the shifting method in \cite{mohassel2017secureml}.
We will further discuss the details in Section \ref{subsec:fixedpoint}.

% Arithmetic sharing part
In the semi-honest 3-server setting, addition and multiplication of the shared values can be easily implemented. 
In order to efficiently compute non-linear functions, we propose the \textit{compute-after-permutation} technique.
% Brief description
To compute the result of a non-linear element-wise function $f$ on shared values $\mathbf x$, $P_0$ and $P_1$ first randomly permute their shares by a common permutation $\pi$ to get $\pi[\mathbf x]$, then send them to $P_2$. 
$P_2$ computes $f(\mathbf x)$ and re-shares it to $P_0$ and $P_1$. 
For element-wise functions, $f(\pi[\mathbf{x}]) = \pi[f(\mathbf{x})]$.
Hence, $P_0$ and $P_1$ can use the inverse permutation to reconstruct the actual shared value of $f(\mathbf x)$. 
% Conclusion 
We will further describe its details in Section \ref{subsec:cap}.

% Neural network
In a fully connected neural network, each layer's computation contains a linear transformation and an activation function. 
The linear transformation (with bias) contains an addition (Add) and a matrix multiplication (MatMul), which can be done trivially under A-SS, as described above.
% Add, Matmul, Activations
As for activation functions, many of them including Sigmoid, ReLU, and Tanh are all element-wise, hence can be computed via the compute-after-permutation technique.
% Other operations
There are also other operations like Transpose that are essential to the computation of neural networks.
Those operations on shared values can be done locally on each party since they are not relevant to actual data values.
% Put together
Hence, after implementing those basic operations in a shared manner, we can build a shared neural network, which takes shared values as input and outputs shared results, piece by piece.
We will describe the details for neural network inference and training in Section \ref{subsec:nn}.

\subsection{Fixed-Point Arithmetic}
\label{subsec:fixedpoint}
% Why this have to be discussed
To maintain safety, A-SS is performed on the integer ring $\mathbb Z_{2^L}$.
Since the computations in machine learning are in float-point format, we have to convert float-point numbers into fixed-point. 
% Existing naive method 
A common solution is to multiply the float-point number by $2^p$~\cite{mohassel2017secureml}. 
This conversion is naturally suitable for addition and subtraction of both normal values and shared values. 
However, for multiplication, the result needs to be truncated via shifting right by $p$ bits.
Considering multiplication of two numbers $x$ and $y$, we have:
$
   \lfloor x y \cdot 2^p\rceil = 
   (\lfloor x \cdot 2^p \rceil \cdot \lfloor y \cdot 2^p \rceil) / 2^p
$ (with a small round-off error),
where $\lfloor \cdot \rceil$ means to cast a real number into integer in ring $\mathbb Z_{2^L}$. 
For simplicity, the operators $+, -$ mentioned below are defined on the ring $\mathbb Z_{2^L}$.
% Negative values
For negative values, we use $2^L - x$ to represent $-x$ where $x\in[1, 2^{L-1}]$.

For multiplication of shared values, the shifting method still can be adopted.
As discussed in \cite{mohassel2017secureml}, the shifting method works for arithmetic shared values except for a very small probability of $\dfrac{1}{2^{L - 1 - 2p}}$ for shifting error. 
Hence, some works~\cite{wagh2019securenn, mishra2020delphi} directly adopted this method.

% Why they are not good
However, when a machine learning model (e,g., a neural network) becomes too complicated, the error rate will not be negligible anymore. 
For example, if we choose $p=20$ and $L=64$, the error rate is $\dfrac{1}{2^{13}}$. 
When performing linear regression inference with an input dimension of 100, there are 100 multiplications for each sample.
Then the error rate of inference for one sample is about $\dfrac{1}{100}$.
Obviously, such a high error rate is not acceptable.

Moreover, the shifting error could be very large and results in accuracy loss for machine learning models.
% Example of truncation error
Here we describe an example of shifting error.
Let $L=64, p=20$, $z$ is a result of multiplication before shifting, and $\langle z\rangle_0 = \langle z\rangle_1 = 2^{63} + 2^{20}$. 
Clearly, the desired result after shifting is 
$\dfrac{(2^{63} + 2^{20})\times 2 \bmod 2^{64}}{2^{20}} = 2$ 
(since it is performed on integer ring $\mathbb Z_{2^{64}}$).
In contrast, the simple shifting method introduced by \cite{mohassel2017secureml} yields a result $\dfrac{2^{63} + 2^{20}}{2^{20}} \times 2 = 2^{43} + 1 +2^{43} + 1 = 2^{44} + 2$. 
We can see that since this error is caused by overflow/underflow, and thus is very large.
In experiments, we also noticed that this error caused the model training to be very unstable, unable to achieve the same performance as centralized plaintext training. 

% Existing works' solutions / our method is not suitable
Some work \cite{demmler2015aby, mohassel2018aby3, riazi2018chameleon} proposed to share conversion protocols to overcome this problem. 
However, since our method does not involve any other sharing types except A-SS, the share conversion is not suitable.

% Our solution
Our solution to this problem is to let each party clip its share so that the signs of their shared values are opposite. 
By doing this, we can avoid the error in the shifting method.
We term it as ShareClip.
% Why?
We prove that when the real value $x$ is within a certain range, this clipping method can prevent all underflow/overflow situations encountered in truncation.

% Proof
\begin{theorem}
If a shared value $x$ on ring $\mathbb Z_{2^L}$ satisfies

$x \in [-2^{L-2}, 2^{L-2})$, and 
$\langle x \rangle_0 \in [-2^{L-2}, 2^{L-2})$, 

then 
$\lfloor \dfrac{\langle x \rangle_0}{2^p} + \dfrac{\langle x \rangle_1}{2^p} \rceil \in [\lfloor \dfrac{x}{2^p} \rceil - 1, \lfloor \dfrac{x}{2^p} \rceil + 1]$.
\end{theorem}

\begin{proof}
Obviously $\langle x\rangle_1 \in [-2^{L-1}, 2^{L-1})$ and no overflow/underflow is encountered when computing $\langle x\rangle_0 + \langle x\rangle_1$. 
Suppose $\dfrac{x}{2^p} = c, \dfrac{\langle x\rangle_0}{2^p} = a,
\dfrac{\langle x\rangle_1}{2^p} = b$, we can easily obtain the result from the fact that if $a + b = c$,  $\lfloor a \rceil + \lfloor b \rceil \in [\lfloor c\rceil - 1, \lfloor c \rceil + 1]$.
\end{proof}

We describe our proposed ShareClip protocol in Algorithm \ref{alg:clip}. 
Its basic idea is that two party interactively shrink their shares when the absolute values of shares are too large. 
Notably, it can be executed during Mul/MatMul of shared values without any extra communication rounds.

\begin{algorithm}[t]
    \caption{ShareClip}
    \label{alg:clip}
    \SetAlgoLined
    \KwIn{Shared value $X$}
    \KwOut{Clipped shared value $Y$}
    $P_0$ finds the indices of elements in $\langle X\rangle_0$ that are greater than $2^{L-2}$ as \textit{indices\_overflow} and the indices of elements that are smaller than $-2^{L-2}$ as \textit{indices\_underflow}, and then sends them to $P_1$\;

    $P_0$ and $P_1$ copy $X$ to $Y$\;

    \For{index in indices\_overflow}
    {
        $P_0$ set $\langle Y\rangle_0[index] \leftarrow \langle Y\rangle_0[index] - 2^{L-2}$\;
        $P_1$ set $\langle Y\rangle_1[index] \leftarrow \langle Y\rangle_1[index] + 2^{L-2}$\;
    }

    \For{index in indices\_underflow}
    {
        $P_0$ set $\langle Y\rangle_0[index] \leftarrow \langle Y\rangle_0[index] + 2^{L-2}$\;
        $P_1$ set $\langle Y\rangle_1[index] \leftarrow \langle Y\rangle_1[index] - 2^{L-2}$.
    }
\end{algorithm} 

\nosection{Choice of $L$ and $p$}
% Choice of L
For ease of implementation, we set $L = 64$, since 64-bit integer operations are widely supported by various libraries. 
% Choice of p
In order to preserve precision for fixed-point computation, we set the precision bits $p = 23$ as suggested in \cite{courbariaux2014lowprecision}. 
Besides, to avoid overflow/underflow during multiplication while using our ShareClip algorithm, all input and intermediate values during computation must be within the range $[-2^{40}, 2^{40})$.

\subsection{Compute-After-Permutation}
\label{subsec:cap}
% Why
Activation functions are essential for general machine learning models such as logistic regression and neural networks. 
In a fully connected neural network, they always come after a linear layer.
% Existing work 
Due to their non-linearity, they cannot be directly computed via A-SS. 
To solve this problem, some methods use GC~\cite{mohassel2017secureml, liu2017minionn}, while others use customized MPC protocols~\cite{mohassel2018aby3, wagh2019securenn}.
% Shortness
However, those methods are usually expensive in implementation, computation and communication. 
Moreover, they usually use approximations to compute non-linear functions, which may cause accuracy loss.

% Our method
In contrast, our proposed \textit{compute-after-permutation} method is quite efficient and easy to implement. 
Since most activation functions are element-wise, we simply randomly permute the input values and let $P_2$ compute the results.
% Correctness and Security
Randomly permuting $n$ elements yields $n!$ possible outcomes. 
Hence, even a few elements can have a huge number of permutations.
During the training and inference of machine learning models, 
input data is always fed in batches, where multiple samples are packed together. 
Besides, the dimension of each input could be large.
Therefore, the inputs of activation functions are always of large size. 
%
% Moreover, at least one linear transformation is applied to the original input data to generate the hidden representation.
%
As a consequence, when $P_2$ gets those permuted hidden representations, it has little chance to guess the original hidden representation, and also little knowledge about the original data.
We will further analyze its security in Section \ref{section:security}.

We formalize the compute-after-permutation technique in Algorithm \ref{alg:cap}, and describe it as follows: 
take neural network for example, assume that $P_0$ and $P_1$ already calculated the shared hidden output $\mathbf{z}$.
% P0 & P1 permute 
First, $P_0$ and $P_1$ use their common Pseudorandom Generator (PRG) to generate a random permutation $\pi$ (line 10).
Then they permute their shares and get $\pi[\langle \mathbf z\rangle_0]$ and $\pi[\langle \mathbf z\rangle_1]$ (line 11). 
After this, they both send the permuted shares to $P_2$.
% P2 compute
When $P_2$ receives these shares, it reconstructs $\pi[\mathbf z]$ by adding them up (line 12). 
Then $P_2$ decodes $\pi[\mathbf z]$ to its corresponding float value and uses ordinary ways to compute the activation $f(\pi[\mathbf z])$ easily (line 13$\sim$14).
After encoding $f(\pi[\mathbf z])$ into fixed point, $P_2$ needs to reshare it to $P_0$ and $P_1$.
% P2 share
To do it, $P_2$ picks a random value $\mathbf r \stackrel{\$}{\leftarrow} \mathbb Z_{2^L}^{\text{Shape}(\mathbf z)}$ 
with the same shape of $\mathbf z$ (line 15),
then sends $\mathbf r$ to $P_0$ and $\mathbf \pi[\mathbf z] - \mathbf r$ to $P_1$ (line 16).
% P0 & P1 permute back
Then $P_0$ and $P_1$ use the inverse permutation $\pi^{-1}$ to get $\pi^{-1}[\mathbf r]$ and $f(\mathbf z) - \pi^{-1}[\mathbf r]$ respectively (line 17). 
Thus, they both get their shares of $f(\mathbf z)$.

\nosection{Enhancing Security via Random Flipping}
% Protecting the prediction
%
Hidden representations are already the transformations of the original data since at least one prior layer is passed.
Hence, the permutation of hidden representation does not leak the value sets of the original data. 
%
% For last layer, it may reflect label
%Why
However, many neural networks still have an activation on the final prediction layer. 
In many tasks, the dimension of the label is just one.
% Examples
For example, when a bank wants to predict the clients' possibility of credit default, the label is 0 or 1 (not default or default),
and when a company wants to predict the sales of products, the label is also one-dimensional and probably between 0 and 1 because normalization is used in the preprocessing.
% It is not safe
In those cases, random permutation only shuffles the predictions within a batch, but the set of the predictions is still preserved.
Hence, $P_2$ can learn the distribution of the predictions in each batch.
Considering that the predictions can be close to the label values in training or inference, the distribution of the predictions is also sensitive and needs to be protected.

% Our solution
In order to prevent this potential information leakage, we propose to add random flipping to the random permutation. 
% Details
Assume $\mathbf z \in \mathbb{R}^B$ is the output of the last layer. 
Before applying the random permutation $\pi$ to it, we first generate a random mask $\mathbf m \stackrel{\$}{\leftarrow} \{0, 1\}^B$. 
Then, for each element in $\mathbf z$, if the corresponding mask is $1$, we change it to its negative. 
Notice that for Sigmoid, we have $\text{Sigmoid}(-x) = 1 - \text{Sigmoid}(x)$, and for Tanh, we have $\text{Tanh}(-x) = - \text{Tanh}(x)$. 
Hence, $P_0$ and $P_1$ can easily get the real value of $f(\mathbf {z})$ from $P_2$'s result. 
The random flipping mechanism is summarized in lines 2$\sim$9 and 18$\sim$25 of Algorithm \ref{alg:cap}.

\begin{algorithm}[t]
    \caption{Compute-after-Permutation(CAP)}
    \label{alg:cap}
    \SetAlgoLined
    \KwIn{
        Shared value $Z$, \newline
        Element-wise function $f \in \{\text{Relu}, \text{Sigmoid}, \text{Tanh}\}$, \newline
        Flipping $\in \{\text{True}, \text{False}\}$ 
    }

    \KwOut{
        Shared element-wise function result $Y = f(Z)$
        }

    $P_0$ and $P_1$ flatten $Z$ to a one-dimensional vector $\mathbf{z}$ \;
    \uIf{Flipping = \upshape $\text{True}$}
    {
        \uIf{\upshape $f \not\in \{ \text{Sigmoid}, \text{Tanh}\}$}{\textbf{return} Error\;}
        
        $P_0$ and $P_1$ generate a random mask with the same size of $\mathbf{z}$: 
        $\mathbf{m} \stackrel{\$}{\leftarrow} \{0, 1\}^{\text{Size}(\mathbf{z})}$ \;
        
        \For{i = 1 to \upshape $\text{Size}(\mathbf{z})$}
        {
            \uIf{$\mathbf{m}_i = 1$}
            {
                $P_0$ and $P_1$ compute $\mathbf{z}_i \leftarrow - \mathbf{z}_i$ by negate their shares\;
            }
        }
    }
    $P_0$ and $P_1$ generate a random permutation $\pi = j_1j_2...j_n$ using their common PRG\;
    
    $P_0$ and $P_1$ permute their shares of $\mathbf{z}$ respectively, generate new shared value $\mathbf{z}'=(z_{j_1}, z_{j_2}, ..., z_{j_n})$\;
    
    $P_0$ and $P_1$ reconstruct $\mathbf{z}'$ to $P_2$\;
    
    $P_2$ gets $\mathbf{z}'$ then decodes it to its float-point value $\mathbf{z}'_f$\; 
    
    $P_2$ computes element-wise function result $\mathbf{y}'_f = f(\mathbf{z}'_f)$ 
    and encodes it to fixed-point value $\mathbf{y}'$\;
    
    $P_2$ generates a random vector $\mathbf{r} \stackrel{\$}{\leftarrow} \mathbb Z_{2^L}^{\text{Size}(\mathbf z)}$
    
    $P_2$ sends $\mathbf{r}$ to $P_0$ and $\mathbf{y'} - \mathbf{r}$ to $P_1$\;
    
    $P_0$ and $P_1$ apply the inverse permutation $\pi^{-1}$ to their shares of $\mathbf{y}'$ and get new shared value $\mathbf{y}$\;
    
    \uIf{Flipping = \upshape $\text{True}$}
    {
        \For{i = 1 to \upshape $\text{Size}(\mathbf{z})$}
        {
            \uIf{$\mathbf{m}_i = 1$}
            {
                \uIf{\upshape$f = \text{Sigmoid}$}
                {
                    $P_0$ and $P_1$ compute $\mathbf{y}_i \leftarrow \text{Sub}(1, \mathbf{y}_i)$\;
                }
                \uIf{\upshape$f = \text{Tanh}$}
                {
                    $P_0$ and $P_1$ compute $\mathbf{y}_i \leftarrow - \mathbf{y}_i$\;
                }
            }
        }
    }
    
    $P_0$ and $P_1$ reshape $\mathbf{y}$ to the shape of $Z$ to get $Y$\;
\end{algorithm}

\nosection{Improving Communication via Common Random Generator}
% Existing work 
To reduce the communication for multiplications, \cite{riazi2018chameleon} used a PRG shared between $P_1$ and $P_2$. 
We also adopt this method for multiplication and further extend this method to our compute-after-permutation technique. 
% Detail
At the offline stage, $P_1$ and $P_2$ set up a common PRG $prg$. 
As mentioned above, after $P_2$ computed the results of element-wise functions, it generates a random vector $\mathbf{r}$ as $P_1$'s share.
In order to reduce communication, $\mathbf{r}$ is generated by $prg$. Because $P_1$ has the same PRG, $P_2$ does not need to send $\mathbf{r}$ to $P_1$.
% The result of reducing communication.
This helps to reduce the network traffic of element-wise functions from $4NL$ bits to $3NL$ bits, where $N$ is the number of elements in the input.
% Comparison
We compare our compute-after-permutation technique with the state-of-the-art MPC protocols on ReLU function in Table \ref{table:cap_comp}. 
% Conclusion
Theoretical result shows that compute-after-permutation requires fewer communication rounds and less network traffic than existing MPC protocols.

\begin{table}[t]\small
    \caption{Comparison of communication rounds and traffics with existing MPC protocols on ReLU function, where $p$ is some prime number and $k$ is the security parameter for GC which is usually more than 128.}
    \label{table:cap_comp}
    \centering
    \begin{tabular}{ccc}
    \toprule
    Protocol & Rounds & Bits \\ 
    \midrule
    Ours & $3$ & $3L$ \\
    SecureNN\cite{wagh2019securenn} & $11$ & $8L\log p + 32L + 2$ \\
    ABY3\cite{mohassel2018aby3} & $6 + \log L$ & $105L$ \\
    Trident\cite{ajith2020trident} & $7$ & $16L + 64$ \\
    GC\cite{rouhani2018deepsecure} & $4$ & $k(3L - 1)$ \\
    \bottomrule
    \end{tabular}
\end{table}

\subsection{Neural Network Inference and Training}
\label{subsec:nn}
% Overview
Based on A-SS and the compute-after-permutation techniques, we can implement both inference and training algorithms for general machine learning models such as fully connected neural networks. 
% Operations
The operations required for neural network inference and training can be divided into three types:

\begin{itemize}
    \item Linear operations: Add, Sub, Mul, and MatMul;
    \item Non-linear element-wise functions: Relu, Sigmoid, and Tanh;
    \item Local transformations: Transpose.
\end{itemize}

We have already implemented linear operations and non-linear element-wise functions by A-SS and the compute-after-permutation technique.
As for the Transpose operator, it can be trivially achieved by $P_0$ and $P_1$ through transposing their shares locally without any communication. 
All these operators have inputs and outputs as secret shared values.
Hence, using those operators, we can implement neural network inference and training algorithms, as described in Algorithm \ref{alg:nn-infer} and Algorithm \ref{alg:nn-backprop}.
Notably, the Sum is performed similarly as Add, and the differentiation of activation function ($\dfrac{\dif a}{\dif x}$) can be the combination of the implemented operators (functions).

\begin{algorithm}[ht]
    \caption{NN-Infer}
    \label{alg:nn-infer}
    \SetAlgoLined
    \KwIn{
        Shared batch data $X$,\newline 
        Shared network parameters (weights, bias, activations)$\{W_i, \mathbf{b}_i, a_i\}$
    }
    \KwOut{Network output $Y$}
    $A_0$ = $X$\;
    \For{i from 0 to num\_layers - 1}
    {
        $Z_{i + 1}\leftarrow$ Add(MatMul($A_i$, $W_i$), $\mathbf{b}_i$)\;
        $A_{i + 1}\leftarrow$ CAP($Z_{i + 1}, a_i, (i = num\_layers - 1)$)\;
    }
    $Y \leftarrow A_{num\_layers}$
\end{algorithm} 

\begin{algorithm}[ht]
    \caption{NN-Backprop}
    \label{alg:nn-backprop}
    \SetAlgoLined
    \KwIn{
        Shared batch data $X$,\newline
        Shared label $Y$,\newline 
        Shared network parameters $\{W_i, \mathbf{b}_i, a_i\},$\newline
        Learning rate $lr$
    }
    \KwOut{Network output $Y$}
    $\hat Y \leftarrow$ NN-Infer($X, \{W_i, \mathbf{b}_i, a_i\}$)\;
    $g_{num\_layers} \leftarrow$ Mul(2, Sub($Y, \hat{Y}$))\;
    \For{i from num\_layers - 1 to 0}
    {
        $g_{i+1} \leftarrow \dfrac{\dif a_{i+1}}{\dif x}(A_{i+1}) \cdot g_{i+1}$\;
        $g_i \leftarrow$ MatMul($g_{i + 1}$, Transpose($W_i$))\;
        $\mathbf{b}_i \leftarrow \mathbf{b}_i - lr\cdot\text{Sum}( g_i, axis=0)$\;
        $W_i \leftarrow W_i - lr \cdot \text{MatMul}(\text{Transpose}(A_i), g_i)$\;

    }
\end{algorithm}

\section{Security Analysis}
\label{section:security}
% Basic setting
We adopt the semi-honest (honest-but-curious) setting in this paper, where each party will follow the protocol but try to learn as much information as possible from his own view~\cite{evans2017pragmatic}. 
In reality, $P_0$ and $P_1$ are data holders while $P_2$ can be a party that both $P_0$ and $P_1$ trust, e.g., a privacy-preserving service provider, a server controlled by government. 
And since in our method, the computation of $P_2$ is simple, it is convenient to put $P_2$ into a TEE (Trusted Execution Environment)~\cite{sabt2015tee} device to enhance security.

% Arithmetic sharing is secure
The security of linear and multiplication operations of A-SS values is proved to be secure under this setting in previous work~\cite{wagh2019securenn} using the universal composability framework~\cite{canetti2001uc}. 
Here, we only need to demonstrate the security of the compute-after-permutation technique.
%

% Empirical
To do this, we will first empirically study the insecurity of directly revealing hidden representations and explain why compute-after-permutation is practically secure.
% Quantitative
Then we will quantitatively analyze the distance correlation between the data obtained by $P_2$ and the original data with or without random permutation. 
We will also derive a formula of expected distance correlation for certain random linear transformations, and an estimation for random permutations.
% We conclude that after performing random permutation on hidden representations, the distance correlation to the original data can usually be lower than reducing the dimension of hidden representations to only 1,
% %
% and verify this conclusion via simulated experiments on different distribution of data.

\subsection{Insecurity of Directly Revealing Hidden Representations}
Since our method is mainly based on the permutation of hidden representations, 
here we demonstrate the insecurity of the methods that directly reveal the hidden representations such as split learning~\cite{vepa2018split}. 
% Their claim
The security guarantee of split learning is the non-reversibility of hidden representations.
For a fully connected layer, the hidden representation (i.e. the output of that layer) can be considered as a random projection of the input. 
For example, the first layer's output is $Y = XW$ (without bias), 
where $X \in \mathbb R^{B\times D}$ is the original data and $W \in \mathbb R^{D\times H}$ is the transformation matrix.
Anyone who gets $Y$ cannot guess the exact value of $X$ and $W$ due to there are infinite choices of $X'$ and $W'$ that can yield the same product $X'W'=XW$.
Thus, it is adopted by existing works like \cite{zhangqiao2019gelunet, vepa2018split, he2020transnet}.

% Why it is not good
However, although the adversary cannot directly reconstruct the original data from hidden representations, there are still potential risks of privacy leakage.
For example, \cite{harry2020exposing} has proven that the methods in \cite{xie2019bayhenn, zhangqiao2019gelunet} may leak information about the model weights when the adversary uses specific inputs and collects the corresponding hidden representations.
% Split-CNN
\cite{abuadbba2020splitcnn} used distance correlation to show that when applying split learning to CNN, the hidden representations are highly correlated with the input data. 
Moreover, the input data may be reconstructed from hidden representations.
% Random projection reconstruction
Older researches like \cite{sang2012reconstructrp} also suggested that it is possible to reconstruct original data from its random projection with some auxiliary information. 
% Utility is p
Besides, the non-reversibility of random projection only protects the original data, but not the utility of data~\cite{bingham2001rp}.
% The utility is preserved
Suppose that company A and company B jointly train a model via split learning, where the training samples are provided by A and the label is provided by B. 
B can secretly collect the hidden representations of the training samples during training.
Then B can either use those hidden representations on another task or give them to someone else without A's permission.

% Our demonstration
% Topic sentence
In order to show the insecurity of revealing hidden representation and the effectiveness of random permutation, 
we propose a simple attack based on the histogram of distances.
The purpose of the attack is to find similar samples according to a certain hidden representation.
Since Johnson-Lindenstrauss lemma~\cite{johnson1984lemma} states that random projection can approximately preserve distances between samples,
it is natural that the histogram before and after random projection can be similar.
Hence, for one sample $\mathbf{y}$ in a projected set $Y = XW$,
the adversary first computes the distances between $\mathbf{y}$ and any other sample $\mathbf{y}'$, 
and then draws the histogram of those distances.
If the adversary has a dataset $X'$ with the same distribution of the original dataset $X$, it then can find the samples in $X'$ which have similar histograms with $\mathbf{y}$. 
Those samples are similar with the original sample corresponding to $\mathbf{y}$.
% Name
We name this simple attack as \textit{histogram attack}.

\begin{figure}[t]
\centering
\includegraphics[width=7cm]{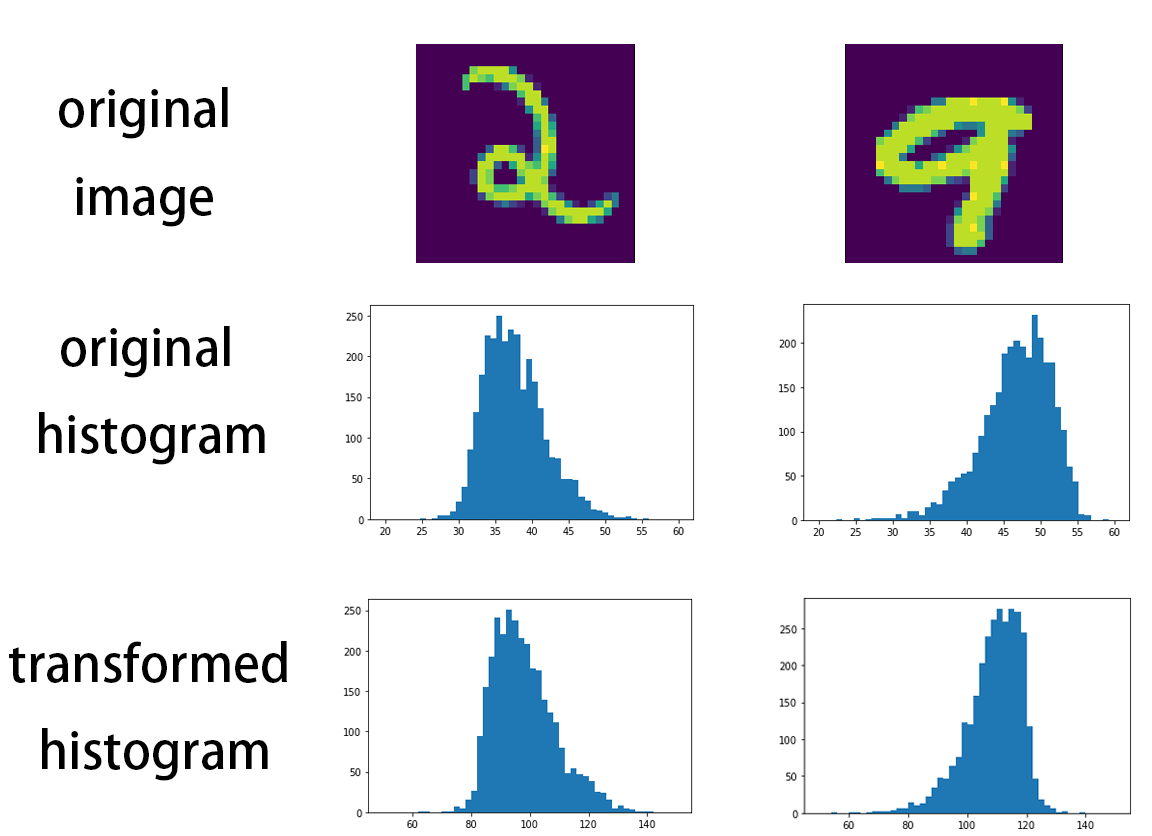}
\caption{The histogram of distances are similar after linear transformation.}
\label{fig:histo}
\end{figure}

\begin{figure}[t]
\includegraphics[width=7.5cm]{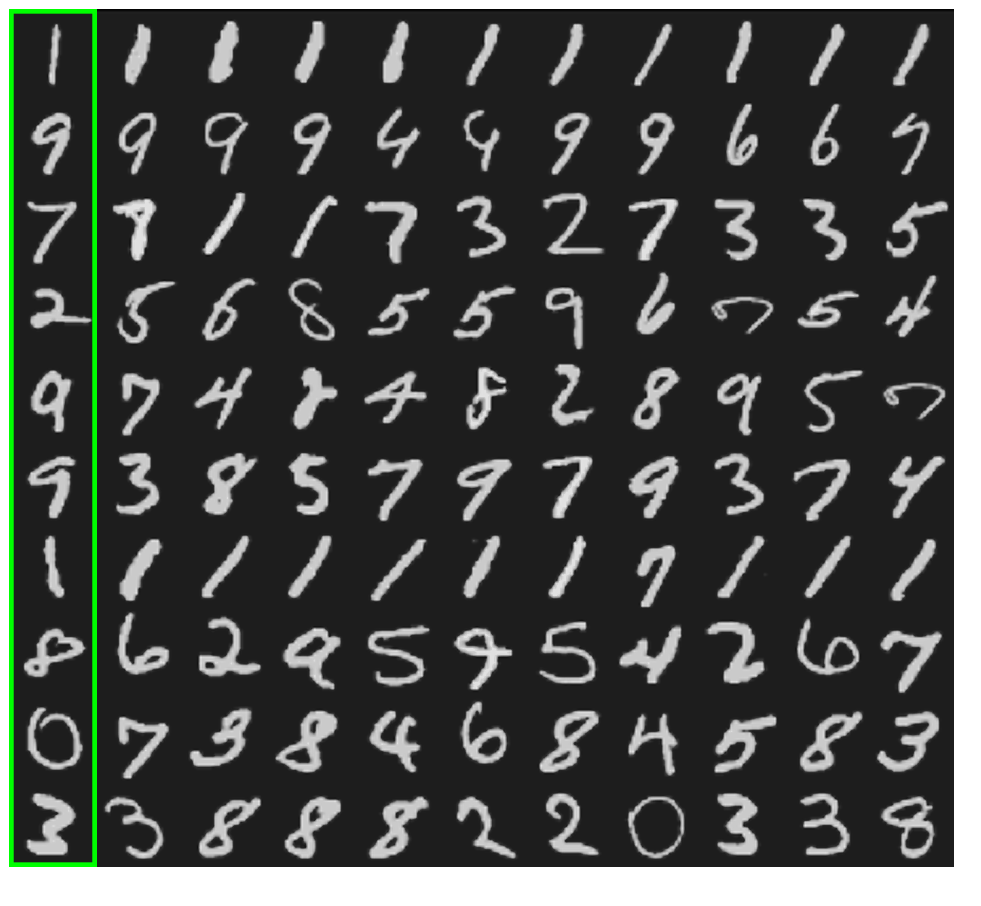}
\caption{By comparing histograms of distances, we can find the similar samples from $X'$ (the most left column is the original sample in $X$, and the other columns are the top-10 similar samples found in $X'$).}
\label{fig:rp}
\end{figure}

% Histogram is similar after transformation
We now use the MNIST~\cite{lecun1998mnist} dataset as an example.
% We can find similar samples from histograms
Assume the adversary has 3,000 images randomly chosen from the MNIST dataset,
along with 128-dimensional hidden representations of 3,000 other images from the dataset.
As shown in Figure \ref{fig:histo}, the histograms of distances before and after random projection are somewhat similar.
The adversary randomly chooses 10 projected vectors and finds their top-10 most similar samples based on the histogram using earth mover's distance~\cite{rubner2000emd}. 
We present the original image and the similar images found via histograms in Figure \ref{fig:rp}.
Obviously, those top-10 images are quite similar to the original one.
For example, for the images of digit 1, almost all of the top-10 similar images are actually of digit 1.

We conclude that directly revealing the hidden representations has two major security concerns:
\begin{itemize}
    \item It preserves the utility of the original data, and may be used multiple times without the permission of data sources.

    \item It preserves the topological structure of the dataset to some extent. 
    If the adversary knows the distribution of the original data, it can find similar samples corresponding to the hidden representation.

\end{itemize}

\subsection{Security of Compute-after-Permutation}
The compute-after-permutation technique is mainly based on random permutation, while for model predictions, random flipping is used.
We demonstrate that random permutation can protect the original data, and the random flipping can protect the model predictions and the label.

\nosection{Random Permutation Protects the Original Data}
Random permutation is a simple technique with strong privacy-preserving power.
Even with a set of very few elements, there are an extremely large number of possible permutations. 
E.g., 10 distinct elements have more than 3 million permutations, while with 20 distinct elements the number of permutations scales to the magnitude of $10^{18}$.
% Hard to guess
Hence, under a mild assumption that the hidden representation of the original data batch has at least a few distinct elements,
it is impossible for the adversary to guess the original hidden representation from its random permutation.

A disadvantage of random permutation is that it preserves the set of elements. 
However, our method is not affected by this disadvantage mainly for two reasons:
\begin{itemize}
    \item In our method, the permutation is performed on the hidden representations instead of the original data, 
    the set of original data values are not exposed.
    \item During the training and inference of machine learning models, the data is always fed in batches. 
    The random permutation is performed on the whole batch of hidden representations.
    Hence, the elements in different samples' hidden representations are mixed together,
    making it hard to extract any individual sample's information.
\end{itemize}

In order to measure the privacy-preserving power of random permuation more precisely, we will quantify the privacy leakage of random permutation and linear transformation measured by distance correlation in the following section \ref{sec:quantitative}. 
By quantitative computation and simulated experiments, 
we demonstrate that applying random permutation on hidden representations usually leaks less privacy than reducing the hidden representations' dimension into 1.
%

% Random Flipping
\nosection{Random Flipping Protects Predictions and Label}
The security of random flipping is based on that no matter what the original prediction is, the adversary will get a value with an equal probability of being positive or negative.
We denote the original predictions by $\mathbf{z}$ and the flipped predictions by $\mathbf{z'}$, and we have:  
\begin{equation}
\begin{split}
& P(z'_i < 0) \\
& = P(z'_i < 0|z_i < 0)P(z_i < 0) + P(z'_i < 0|z_i \ge 0)P(z_i \ge 0) \\
& = \dfrac{1}{2}P(z_i < 0) + \dfrac{1}{2}P(z_i\ge 0) = \dfrac{1}{2} = P(z'_i \ge 0).
\end{split}
\end{equation}

Hence, no matter what the original predictions are, $P_2$ only receives a batch of values with equal probabilities of negative or positive. 
When the label is binary, no information about the label is leaked. 
When the label is continuous, the scales of the predictions are leaked. 
However, since the predictions do not exactly match the label and the values are permuted, we consider those flipped values are safe to reveal.

\vspace{\baselineskip}

% Conclusion
As discussed above, with the random permutation performed on hidden representation, 
the adversary cannot reconstruct the original hidden representation and very little information about the original data is leaked.
Also, the random flipping preserves the privacy for model predictions and the label data.
Therefore, we conclude that our compute-after-permutation technique is practically secure.

\subsection{Quantitative Analysis on Distance Correlation}
\label{sec:quantitative}
To better illustrate the security of our compute-after-permutation method, 
we quantify the privacy leakage by distance correlation.
First, we derive a formula of expected distance correlation for random linear transformation. 
Then we demonstrate that random permuted vector can be viewed as the combination of element-wise mean of the vector and an almost-random noise.
Based on this, we conclude that random permutation usually preserves more privacy than compressing data samples to only one dimension.
We also conduct experiments on 4 simulated data distributions to verify our conclusion.

In the following of this section, 
we use $|\mathbf x|$ to denote the euclidean norm of the vector $\mathbf x$, 
upper case letters such as $X$ to denote the corresponding random variable, 
and $\mathbf x_i$ to denote the $i$'th component of the vector.
\\

\subsubsection{Linear Transformation}
\begin{theorem}[DCOR for linear transformations]
\label{theorem:linear-dcor}

Suppose $X\in\mathbb R^n$ is an arbitrary random vector, 
% Linear transformation
and let $Y = AX$, where $A\in \mathbb R^{n\times d}$ is the transformation matrix 
% Rotation-Invariant
% and $A$ is drawn from some distribution 
that satisfies the probability of $P(A) = P(AT)$ for any orthogonal matrix $T$ (i.e., P is a rotation-invariant distribution), 
then \\

$\E\limits_A [Dcor(X, Y)] = \sqrt{\dfrac{a^2(S_1 + S_2 - 2S_3)}{a^2S_1 + b^2S_2 - 2S'_3}}$,
where \\ \\

$S_1 = \E |X-X'|^2$, 
$S_2 = \left[\E |X-X'|\right]^2$,
$S_3 = \E |X-X'||X-X''|$,
$S'_3 = \E\limits_A |X-X'||AX - AX''| = \E C(X-X',X-X'')$,
$X', X''$ are identical independent distributions of $X$,

and 

$a = \E\limits_A\dfrac{|A\mathbf u|}{|\mathbf u|}, b = \sqrt{\E\limits_A\dfrac{|A\mathbf u|^2}{|\mathbf u|^2}}$,\\

$C(\mathbf x, \mathbf y) = \E\limits_A[|g_A(\theta\langle\mathbf x, \mathbf y\rangle)||\mathbf x||\mathbf y|]$,

$g_A(\theta) = \E\limits_A[\dfrac{|A\mathbf x||A\mathbf y|}{|\mathbf x||\mathbf y|}]$
where $\mathbf x, \mathbf y$ denote arbitrary vectors and $\theta$ denotes the angle between them.

\end{theorem}
\begin{proof}
The proof yields directly from the Brownian distance covariance \cite{szekely2009browniandc} formula and the rotation-invariant property of $A$.
\end{proof}

\begin{corollary}
\label{corollary:g_theta}
The function $g_A(\theta)$ is monotonic decreasing for $\theta \in [0, \pi/2]$.
\end{corollary}
\begin{proof}
Due to the rotation-invariant property of $A$, it is sufficient to assume 
$\mathbf x = \begin{bmatrix}\cos\theta/2 \\ \sin\theta/2 \\ \mathbf 0\end{bmatrix}, 
 \mathbf y = \begin{bmatrix}\cos\theta/2 \\ \sin-\theta/2 \\ \mathbf 0\end{bmatrix}$,
then we have $g_A(\theta) = \E\limits_A{|A\mathbf x||A\mathbf y|}$.

By rotating $A$ in the plane spanned by the first two axis, we have 
$AT = A \begin{bmatrix}\cos\alpha & \sin\alpha & O \\ -\sin\alpha & \cos\alpha & O \\ O & O & E  \end{bmatrix}$.
Then we have:
\begin{equation}
\begin{split}
       & \E\limits_A [|Ax||Ay|] = \E\limits_A[|ATx||ATy|] = \E\limits_{A}\E\limits_T[|ATx||ATy|] \\
       =& \E\limits_A \E\limits_\alpha
           \left[\left|\cos(\alpha - \theta/2) \mathbf a_0 - \sin(\alpha - \theta/2)\mathbf a_1\right| \right.\\
         &\hspace{18pt}\left.\cdot\left|\cos(\alpha + \theta/2)\mathbf a_0  - \sin(\alpha + \theta/2)\mathbf a_1\right|\right]
\end{split}
\end{equation},
where $\mathbf a_0$ and $\mathbf a_1$ are first two columns of $A$.

By taking its derivation on $\theta$, we have:

\begin{equation}
\begin{split}
\label{eq:dif_ET}
    & \dfrac{\dif \E_T |AT\mathbf x||AT\mathbf y|}{\dif \theta} = -(|\mathbf a_0|^2 + |\mathbf a_1|^2) \sin\theta \cdot \\
    & \int\limits_{\alpha=0}^{2\pi}\text{sign}\left[ \cos2\alpha(|\mathbf a_0|^2 - |\mathbf a_1|^2) - \sin2\alpha (\mathbf a_0 \cdot \mathbf a_1) \right. \\
    & \hspace{40pt} + \left. (|\mathbf a_0|^2 + |\mathbf a_1|^2)\cos\theta\right] \dif \alpha.
\end{split}
\end{equation}

The first two terms inside the sign function always integral to zero, 
and the third term is a constant positive term.
Hence Eq.~\eqref{eq:dif_ET} is non-positive when $\theta \in [0, \pi/2)$, regardless of $A$.
\end{proof}

The above analysis shows that the expected distance correlation between the original data and random linear transformed data is affected by the distribution of the original data. 
Since the coefficients $a, b$ in Theorem \ref{theorem:linear-dcor} are constant, 
the expected distance correlation mainly relies on term $S'_3$, 
which is a function of angles between different data points. Larger angle leads to smaller distance correlation.

Intuitively, larger angle means data points are distributed more randomly on each dimension, and smaller angle indicates data points are concentrated on some subspaces. 
For example, if all data points are distributed very near to a line, then all angles between $\mathbf x - \mathbf x'$ and $\mathbf x - \mathbf x''$ ($\mathbf x, \mathbf x', \mathbf x''$ are three data points) are near to 0, which leads to a large distance correlation.

For neural networks, the transformation matrix $A\in \mathbb R^{d\times h}$ is initialized with normal distribution $\mathcal N(0, \sigma^2)$, 
which satisfies the rotation-invariant property. 
In this case, we have $a = \sigma\sqrt{2}\dfrac{\Gamma((h + 1)/2)}{\Gamma(h / 2)}$, $b = \sigma\sqrt{h}$,
and $g_A(\theta)$ = $\mathlarger{\int}\limits_{\mathbf x,\mathbf y\in \mathbb R^d}
    e^{-\dfrac{|\mathbf x|^2 + |\mathbf y|^2}{2\sigma ^2}}|\mathbf x||\cos\theta \mathbf x + \sin\theta \mathbf y|\dif \mathbf x \dif \mathbf y$.

\subsubsection{Random Permutation}

% Property of random permuted vector
For a vector $\mathbf x \in \mathbb R^n$, we write its random permutation as $\pi[\mathbf x]$ which has the following properties:

\begin{itemize}
    \item $\E\limits_\pi \pi[\mathbf x] = [\E \mathbf x, ..., \E \mathbf x] = M(\mathbf x)$, 
    where $\E \mathbf x = \dfrac1n\sum\limits_{i=1}^n \mathbf x_i$.
    
    \item $(\pi[\mathbf x] - \E\limits_\pi \pi[\mathbf x])^T \mathbf 1 = 0$
    for any $\pi$.
    
    \item $\left|\pi[\mathbf x] - \E\limits_\pi \pi[\mathbf x]\right| = \left|\mathbf x - \E\limits_\pi \pi[\mathbf x]\right|$
    for any $\pi$.
\end{itemize}

In other words, the permuted vector can be viewed as a sum of the element-wise mean vector $M(\mathbf x)$ and an error vector $\mathbf e_x = \pi[\mathbf x] - \E\limits_\pi \pi[\mathbf x]$. 
The error vector is distributed on a $(n-1)$-sphere on a hyperplane orthogonal to $\mathbf 1$ centered at origin $\mathbf 0$.
We further illustrate that this error vector is approximately uniformly distributed on that sphere.

% error vectors are uniformly distributed (almost)
\begin{theorem}
\label{th:error-vec-orthogonal}
The error vector's projection on any unit vector on the hyperplane $\mathbf y^T \mathbf 1 = 0$ 
has a mean $0$ and a variance of \normalfont $\approx \dfrac1n |\mathbf e_x|^2$.
\end{theorem}

\begin{proof}
The mean of $\mathbf e_x \cdot \mathbf y$ can be computed as follows:
{\parindent0pt
$\E[\mathbf e_x \cdot \mathbf y] = \E\limits_{\pi_x} \sum\limits_{i=1}^n\pi_x [ \mathbf e_x ]_i \mathbf y_i
 = \sum\limits_{i=1}^n \E\limits_{\pi_x} \pi_x[ \mathbf e_x ]_i\cdot \mathbf y_i  = \E\limits_{\pi_x} 0 \cdot \mathbf y_i = 0$.
}

For the calculation of variance, we have: 
\begin{equation}
\label{eq:var-of-error-vector}
    \text{Var}[\mathbf e_x \cdot \mathbf y] = \E\limits_{\pi_x} \sum\limits_{i,j=1}^n 
\pi_x [ \mathbf e_x]_i \mathbf y_i \pi_x [ \mathbf e_x]_j \mathbf y_j.
\end{equation}

The term inside sum can be divided into two cases:

\begin{itemize}
    \item $i = j$: In this case, 
    $\E\limits_{\pi_x} \pi_x [ \mathbf e_x]_i \mathbf y_i \pi_x [ \mathbf e_x]_j \mathbf y_j
    = \E\limits_{\pi_x} \pi_x [ \mathbf e_x]_i^2 \mathbf y_i^2 \\ = \dfrac{|\mathbf e_x|^2}{n}\mathbf y_i^2$.
    
    \item $i\ne j$: In this case,

    $\E\limits_{\pi_x} \pi_x [ \mathbf e_x]_i \mathbf y_i \pi_x [ \mathbf e_x]_j \mathbf y_j
    =\E\limits_{\pi_x} \pi_x [ \mathbf e_x]_i \pi_x [ \mathbf e_x]_j \mathbf y_i \mathbf y_j
    $. 
    
    Observed that $\E\limits_{\pi_x} \pi_x [ \mathbf e_x]_i \pi_x [ \mathbf e_x]_j = \dfrac1{n^2}\sum\limits_{i\ne j}(\mathbf e_x)_i (\mathbf e_x)_j
    \\ = \dfrac1{n^2}\sum\limits_{i} -(\mathbf e_x)_i^2 =- \dfrac1{n^2}|\mathbf e_x|^2$, we have:
    
    $\E\limits_{\pi_x} \pi_x [ \mathbf e_x]_i \mathbf  y_i \pi_x [ \mathbf e_x]_j \mathbf y_j = 
    - \dfrac1{n^2}|\mathbf e_x|^2\mathbf y_i \mathbf y_j$.
\end{itemize}

Then by replacing those terms in Eq.~\eqref{eq:var-of-error-vector} and 
notice that $\sum_{i\ne j} \mathbf y_i\mathbf y_j = -|\mathbf y|^2$, $|\mathbf y| = 1$, we have:

$\text{Var}[\mathbf e_x \cdot \mathbf y] = (\dfrac1n +\dfrac{1}{n^2}) |\mathbf e_x|^2 \approx \dfrac1n|\mathbf e_x|^2$.

\end{proof}

Theorem \ref{th:error-vec-orthogonal} implies that the error vector tends to be uniformly distributed on the $(n-1)$-sphere. 
For any unit vector $\mathbf y$ on that sphere, the variance of the inner product is the same and converges to 0 as the dimension $n$ increases.

% We are working on hidden representations
In our case, let $Y$ be a hidden representation of a sample obtained by random linear transformation of the original data, i.e., $Y = AX + \mathbf b$. 
As long as $A$ and $\mathbf b$ follow normal distribution (or any distribution that has a zero mean), 
we have: $\E Y \approx 0, |M(Y)|^2 \approx \dfrac{1}{n}|Y|^2$ and $|E_Y| \approx |Y|$. 
I.e., the magnitude of $Y$'s error vector $E_Y$ is significantly larger than $Y$'s element-wise mean $M(Y)$.
Then denoting $V$ as the distance covariance function, it is appropriate to assume:

\begin{itemize}
    \item $V(X, E) \approx 0$ and $V(M(Y), E) \approx 0$, 
    i.e., the error vector of $Y$'s permutation is nearly independent with $Y$'s element-wise mean or $X$.
    \item $V(\pi[Y]) = V(M(Y) + E_Y) > V(M(Y))$, 
    i.e., the distance variance of $\pi[Y]$ is smaller than the distance variance of $M(Y)$ due to the magnitude of error vector is large.
\end{itemize}

Under those assumptions and by the property of distance covariance, we have:
\begin{equation}
\begin{split}
    \E\limits_\pi \text{Dcor}(X, \pi[Y])& = V^2(X, \pi[Y])/\sqrt{V^2(X)V^2(\pi[Y])} \\
    &\lesssim V^2(X, M(Y))/\sqrt{V^2(X)V^2(M(Y))} \\
    & = \text{Dcor}(X, M(Y)).
\end{split}
\end{equation}

Since $Y$ is a random projection of $X$, then we have:

\begin{equation}
\begin{split}
\E\limits_{\pi,A}\text{Dcor}(X, \pi[AX])& \lesssim \E\limits_{\pi,A}\text{Dcor}(X, M(AX))
\\& = \E\limits_{B\in\mathbb R^{n\times 1}}\text{Dcor}(X, BX),
 \end{split}
\end{equation}
where both $A$ and $B$ follow random normal distribution.

By now, we can conclude that in the sense of distance correlation, 
applying random permutation on hidden representations usually leaks less information than reducing the hidden representation to only one dimension.

\subsubsection{Simulated Experiment}

In order to verify the above analysis, we conduct simulated experiments on the following four kinds of data distributions whose dimensions are all 100. 
\begin{itemize}
    \item \textit{Normal distribution}. Each data value is drawn from $\mathcal N(0, 1)$ independently.
    \item \textit{Uniform distribution}. Each data value is drawn from $\mathcal U(0, 1)$ independently.
    \item \textit{Sparse distribution}, Each data value has a probability of 0.1 to be 1 and otherwise being 0.
    \item \textit{Subspace distribution}, Each data sample is distributed near to a 20-dimensional subspace and with a error drawn from $\mathcal N(0, 1)$.
    Each data sample can be represented by $X=AH+E$, where $H \in \mathbb R^{20}$ and $H_i \sim \mathcal N(0, 1)$, 
    $A \in \mathbb R^{20\times100}, A_{i,j}\sim \mathcal N(0, \dfrac1{20^2})$, and $E\in \mathbb R^{100}, E_{i} \sim \mathcal N(0, 0.1)$.
\end{itemize}

% Simulated experiments
% Settings
For each distribution, we simulate 10,000 samples. 
For random linear transformation (labeled as random projected), we use Theorem \ref{theorem:linear-dcor} to accurately calculate the expected distance correlation. 
And for random permutation, we use Brownian distance covariance \cite{szekely2009browniandc} to estimate the distance correlation using 10,000 repeated samplings in order to reduce estimation error.

\begin{figure}[t]
\centering
    \begin{subfigure}[b]{0.23\textwidth}
    \centering
    \includegraphics[width=\textwidth]{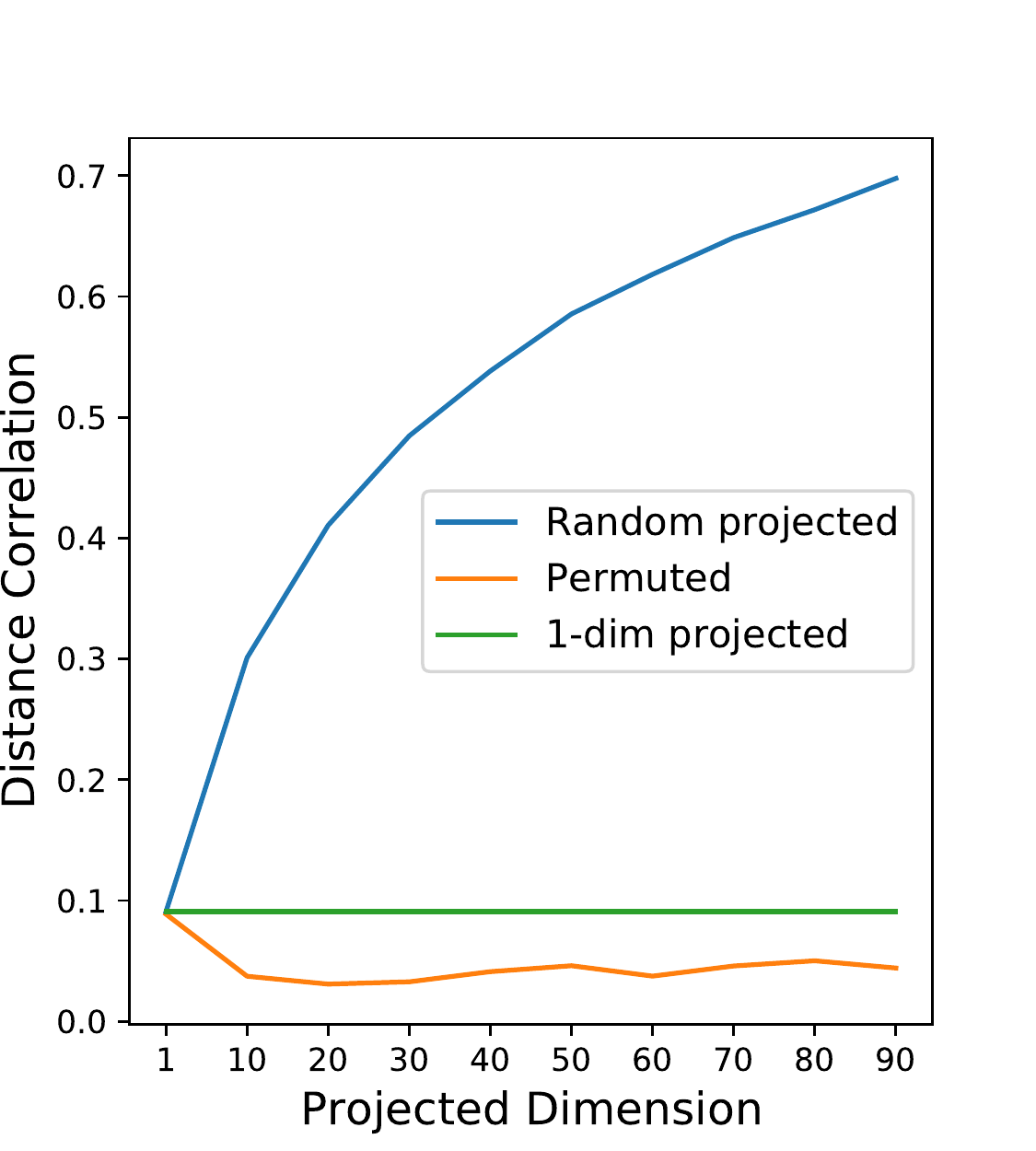}
    \caption{Normal}
    \end{subfigure}
    \begin{subfigure}[b]{0.23\textwidth}
    \centering
    \includegraphics[width=\textwidth]{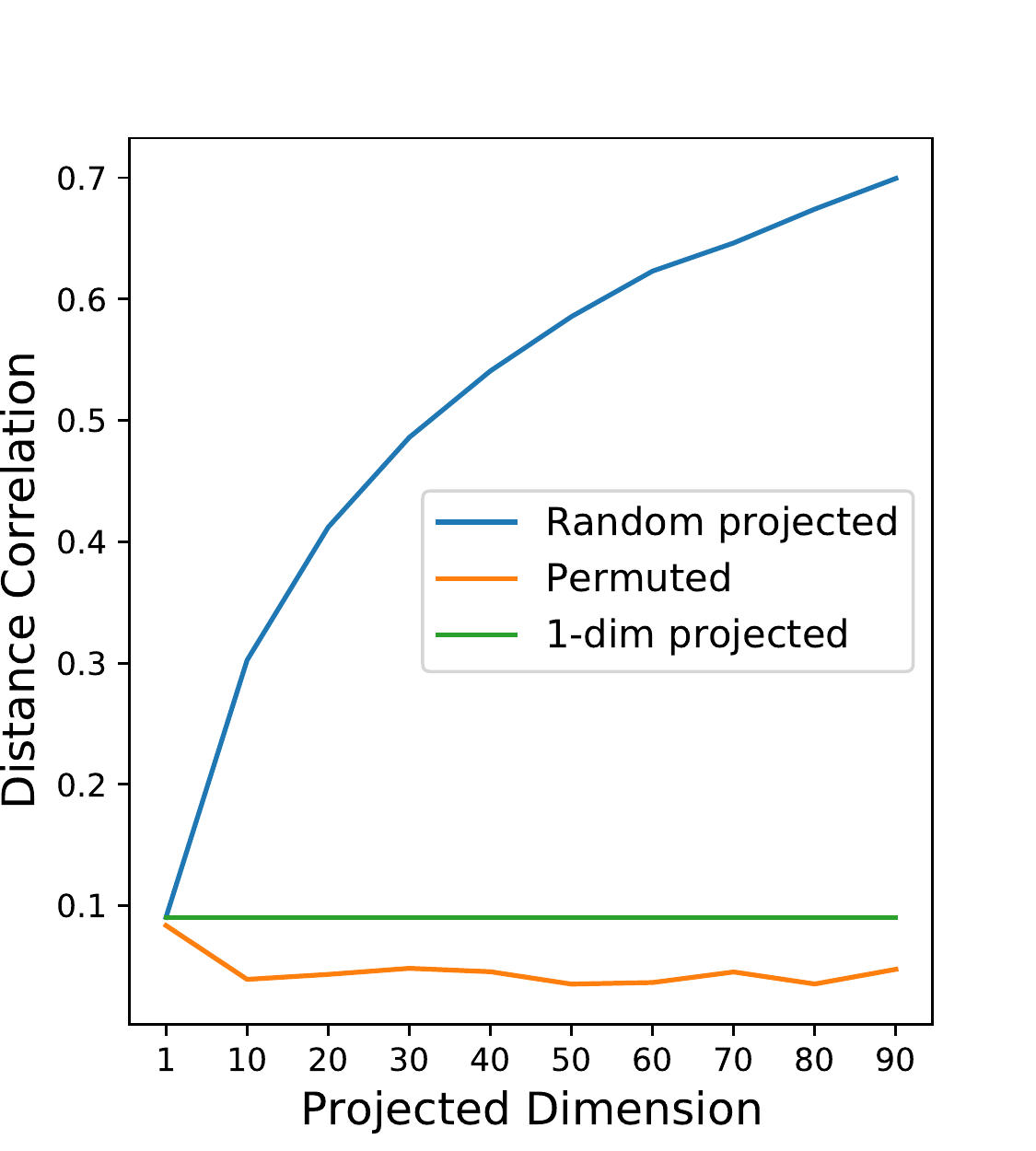}
    \caption{Uniform}
    \end{subfigure}
    
    \begin{subfigure}[b]{0.23\textwidth}
    \centering
    \includegraphics[width=\textwidth]{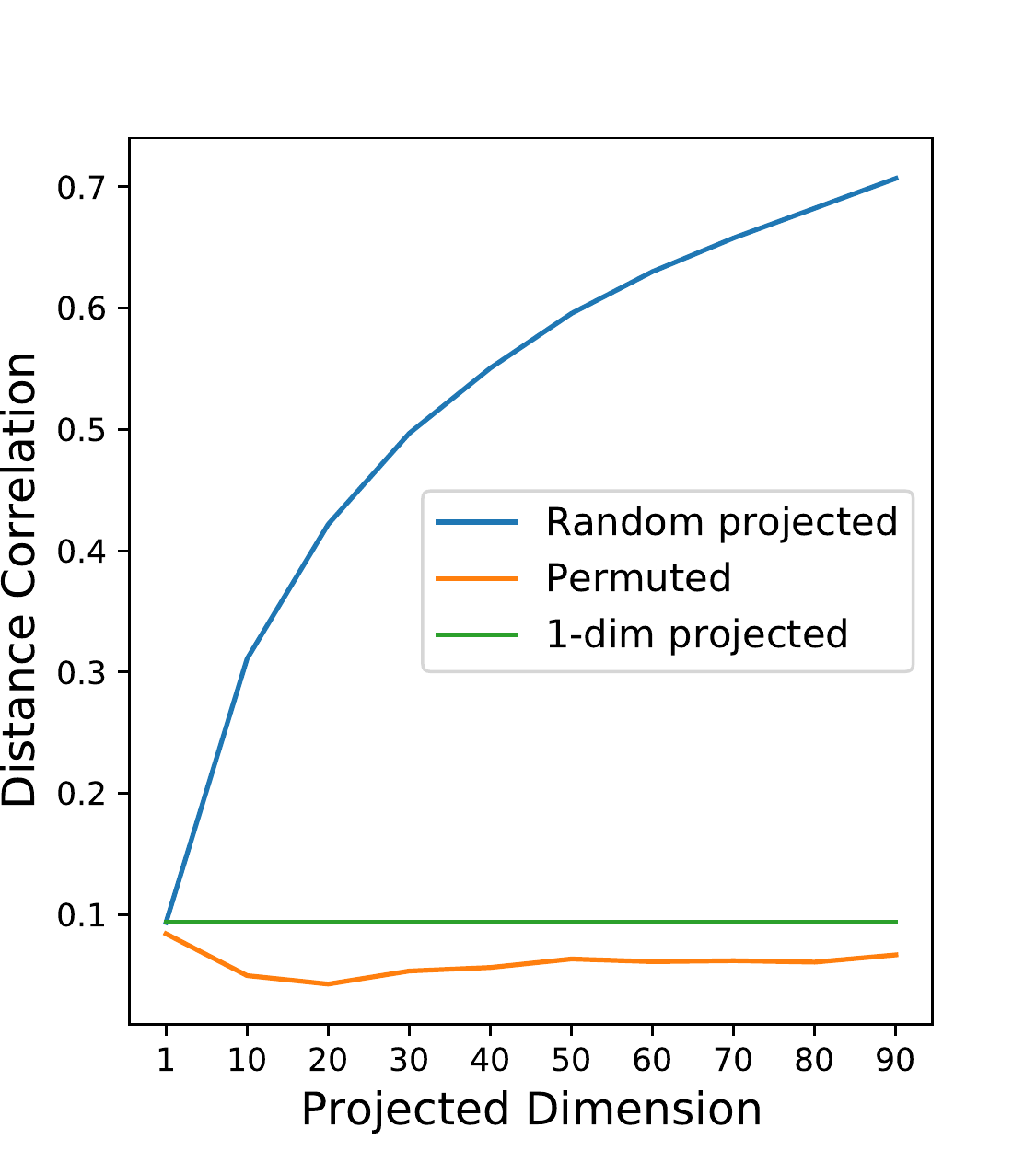}
    \caption{Sparse}
    \end{subfigure}
    \begin{subfigure}[b]{0.23\textwidth}
    \centering
    \includegraphics[width=\textwidth]{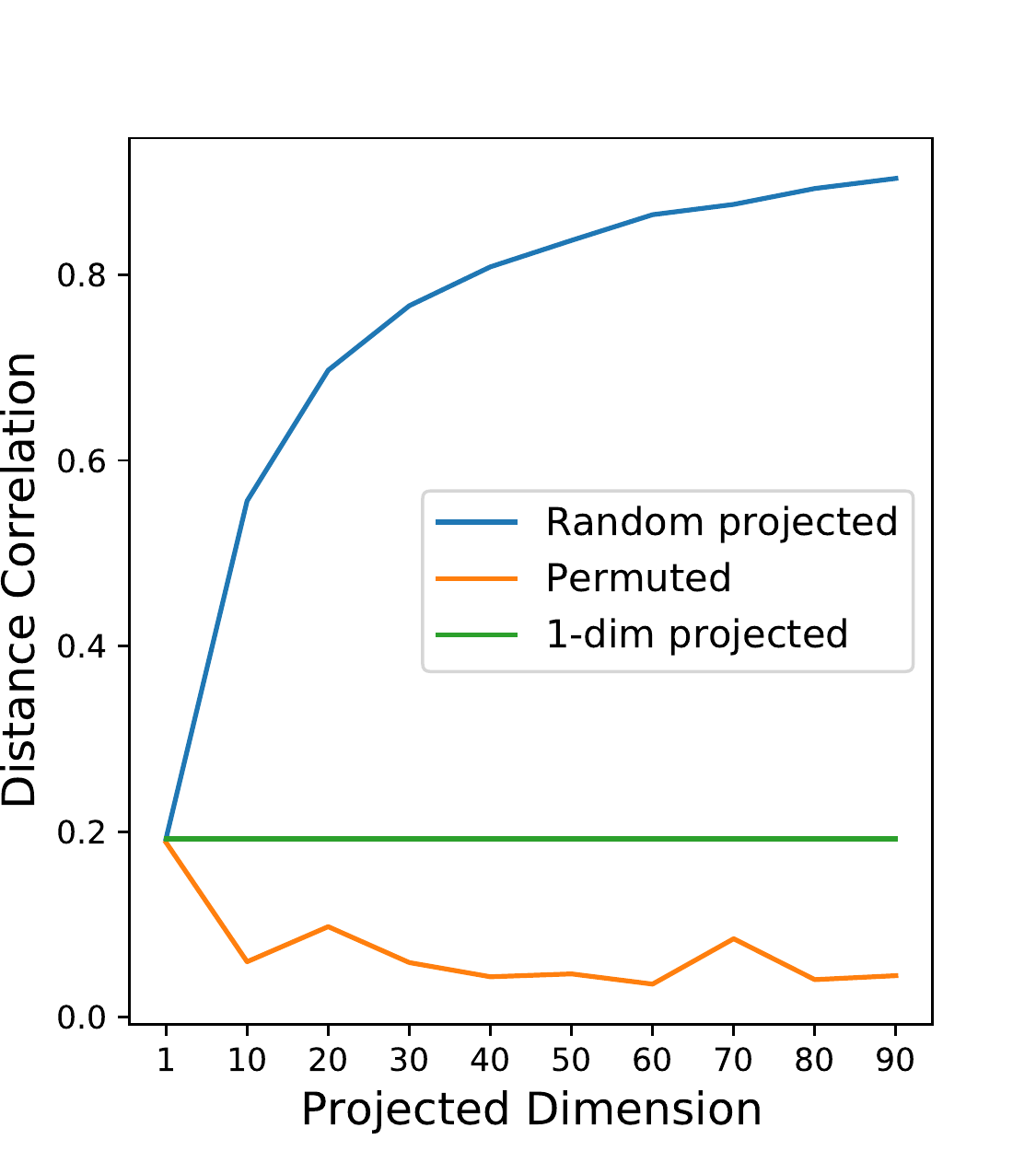}
    \caption{Subspace}
    \end{subfigure}

\caption{Simulated experiments on different distribution. 
Blue: Random linear transformation; Green: random linear transformation to 1 dimension; Orange: Random permutation}
\label{fig:dcors}
\end{figure}

% Results
The results in Figure \ref{fig:dcors} support our analysis. The distance correlation of permuted data is constantly smaller than the distance correlation of 1-dimensional transformed data. 
The results also show that the distance correlation of linear transformation is strongly affected by the distribution of the original data.
In the subspace case, the distance correlation is significantly higher than other distributions.
However, after applying random permutations on hidden representations, the distance correlation drops to a level below 0.1.
This shows random permutation is more resilient when the original data's distribution is special.

\section{Experiments}

In this section, we conduct experiments on both simulated data and real-world datasets.
We demonstrate the efficiency of our method via benchmarking our method on Logistic Regression (LR) and Deep Neural Network (DNN) models, and compare with the state-of-the-art cryptographic methods.
We also demonstrate the security of our method via computing distance correlation and simulating histogram attack on the leaked data.

\subsection{Settings}
% Code
Our implementation is written in Python, and we use NumPy library for both 64-bit integer and float-point computations. 
% Environment Setting
We conduct our experiments on a server equipped with a 16-core Intel Xeon CPU and 64Gb RAM.
We simulate the WAN setting using Linux's tc command. 
The bandwidth is set to 80Mbps and the round trip latency is set to 40ms. 
% Dataset
For all experiments, the data is shared between $P_0$ and $P_1$ before testing.

\subsection{Benchmarks}
We benchmark the running time and network traffic for logistic regression and neural network models using our proposed method, and compare them with ABY3~\cite{mohassel2018aby3} and SecureNN~\cite{wagh2019securenn}. 
We choose the open-source library rosetta\footnote{https://github.com/LatticeX-Foundation/Rosetta} and tf-encrypted\footnote{https://github.com/tf-encrypted/tf-encrypted} to implement SecureNN and ABY3 respectively. 
We run all the benchmarks for 10 times and report the average result. 
To measure the network traffic, 
% measuring details
we record all the network traffics of our method and
use the tshark command to monitor the network traffics for ABY3 and SecureNN. 
We exclude the traffic for TCP packet header via subtracting $64 \cdot N_{packets}$ from the originally recorded bytes.
Since we only benchmark the running time and network traffic, we use random data as the input of the models.

% logistic regression
\nosection{Logistic Regression}
We benchmark logistic regression model with input dimension in \{100, 1,000\} and batch size in \{64, 128\}, for both model training and inference.
The benchmark results for logistic regression model are shown in Table \ref{table:benchmark-lr-time} and Table~\ref{table:benchmark-lr-traffic}. 
Compared with the best results of other methods, our method is about 2$\sim$4 times faster for model inference and training. 
As for network traffic, our method has a reduction of about 35\%$\sim$55\% for model inference and training when the input dimension is 100,
and is slightly higher than ABY3 in the case of dimension 1000.

\begin{table}[t]
\caption{Runing time (s) for LR training/inference.}
\label{table:benchmark-lr-time}
\begin{tabular}{cccccc}
\hline
\multicolumn{1}{l}{Dim}    & \multicolumn{2}{l}{batch size} & Ours           & SecureNN       & ABY3  \\ \hline
\multirow{4}{*}{100}    & \multirow{2}{*}{64}            & infer & \textbf{0.099} & 0.219    & 0.5   \\ \cline{3-6} 
                          &                                & train & \textbf{0.279} & 0.348    & 0.534 \\ \cline{2-6} 
                          & \multirow{2}{*}{128}           & infer & \textbf{0.108} & 0.228    & 0.5   \\ \cline{3-6} 
                          &                                & train & \textbf{0.281} & 0.367    & 0.539 \\ \hline
\multirow{4}{*}{1000}   & \multirow{2}{*}{64}            & infer & \textbf{0.132} & 0.358    & 0.511 \\ \cline{3-6} 
                          &                                & train & \textbf{0.294} & 0.698    & 0.831 \\ \cline{2-6} 
                          & \multirow{2}{*}{128}           & infer & \textbf{0.114} & 0.558    & 0.513 \\ \cline{3-6} 
                          &                                & train & \textbf{0.334} & 1.202    & 0.837 \\ \hline
\end{tabular}
\end{table}

\begin{table}[t]
\caption{Network traffic (Mb) for LR training/inference.}
\label{table:benchmark-lr-traffic}
\begin{tabular}{ccllll}
\hline
\multicolumn{1}{l}{Dim}       & \multicolumn{2}{l}{batch size} & Ours          & SecureNN & ABY3 \\ \hline
\multirow{4}{*}{100}  & \multirow{2}{*}{64}            & infer & \textbf{0.103} & 0.226    & 0.372          \\ \cline{3-6} 
                        &                                & train & \textbf{0.209} & 0.391    & 0.385          \\ \cline{2-6} 
                        & \multirow{2}{*}{128}           & infer & \textbf{0.202} & 0.448    & 0.624          \\ \cline{3-6} 
                        &                                & train & \textbf{0.413} & 0.775    & 0.639          \\ \hline
\multirow{4}{*}{1000} & \multirow{2}{*}{64}            & infer & \textbf{0.996} & 1.072    & 1.369          \\ \cline{3-6} 
                        &                               & train     & 1.988      & 2.581   & \textbf{1.678}          \\ \cline{2-6} 
                        & \multirow{2}{*}{128}           & infer     & 1.975      & 2.134    & \textbf{1.884} \\ \cline{3-6} 
                        &                                & train     & 3.949         & 5.121    & \textbf{3.225} \\ \hline
\end{tabular}
\end{table}

\nosection{Neural Networks}
We also benchmark two fully connected neural networks DNN1 and DNN2 in Table \ref{table:benchmark-dnn-time} and Table \ref{table:benchmark-dnn-traffic}.
DNN1 is a 3-layer fully connected neural network with an input dimension of 100 and a hidden dimension 50, while DNN2 has an input dimension of 1,000 and a hidden dimension of 500.
Compared with logistic regression models, our method has more advantage on neural networks. 
Compared with the neural networks implemented by SecureNN and ABY3, the speedup of our model against model inference/training is about 1.5×$\sim$5.5×, and the reduction of network traffic is around 38\%$\sim$80\%.

\begin{table}[t]
\caption{Running time (s) for DNN inference/training, where DNN1's architecture is 100-50-relu-1-sigmoid and DNN2's architecture is 1000-500-relu-1-sigmoid.}
\label{table:benchmark-dnn-time}
\begin{tabular}{cccccc}
\hline
                      & \multicolumn{2}{c}{batch size} & Ours           & SecureNN & ABY3   \\ \hline
\multirow{4}{*}{DNN1} & \multirow{2}{*}{64}            & infer & \textbf{0.187} & 0.682    & 0.75           \\ \cline{3-6} 
                      &                                & train & \textbf{0.54}  & 1.292    & 0.883          \\ \cline{2-6} 
                      & \multirow{2}{*}{128}           & infer & \textbf{0.198} & 1.097    & 1.776          \\ \cline{3-6} 
                      &                                & train & \textbf{0.589} & 2.083    & 0.916          \\ \hline
\multirow{4}{*}{DNN2} & \multirow{2}{*}{64}            & infer & \textbf{1.047} & 5.672    & 1.862          \\ \cline{3-6} 
                      &                                & train & \textbf{1.864} & 12.109   & 3.236          \\ \cline{2-6} 
                      & \multirow{2}{*}{128}           & infer & \textbf{1.262} & 9.875    & 3.645          \\ \cline{3-6} 
                      &                                & train & \textbf{2.577} & 20.002   & 5.137          \\ \hline
\end{tabular}
\end{table}

\begin{table}[t]
\caption{Total network traffic (Mb) for DNN inference/training.}
\label{table:benchmark-dnn-traffic}
\begin{tabular}{cccccc}
\hline
\multicolumn{1}{l}{}  & \multicolumn{2}{l}{batch size} & Ours           & SecureNN & ABY3  \\ \hline
\multirow{4}{*}{DNN1} & \multirow{2}{*}{64}            & infer & \textbf{0.39}  & 1.984    & 1.895           \\ \cline{3-6} 
                      &                                & train & \textbf{0.78}  & 4.05     & 2.208           \\ \cline{2-6} 
                      & \multirow{2}{*}{128}           & infer & \textbf{0.7}   & 3.891    & 3.629           \\ \cline{3-6} 
                      &                                & train & \textbf{1.38}  & 7.902    & 4.099           \\ \hline
\multirow{4}{*}{DNN2} & \multirow{2}{*}{64}            & infer & \textbf{10.69} & 25.175   & 17.16           \\ \cline{3-6} 
                      &                                & train & \textbf{17.97} & 55.979   & 35.196          \\ \cline{2-6} 
                      & \multirow{2}{*}{128}           & infer & \textbf{12.54} & 43.079   & 36.094          \\ \cline{3-6} 
                      &                                & train & \textbf{24.84} & 93.19    & 47.71           \\ \hline
\end{tabular}
\end{table}

\begin{figure}[t]
\centering
    \begin{subfigure}[b]{0.4\textwidth}
    \centering
    \includegraphics[width=\textwidth]{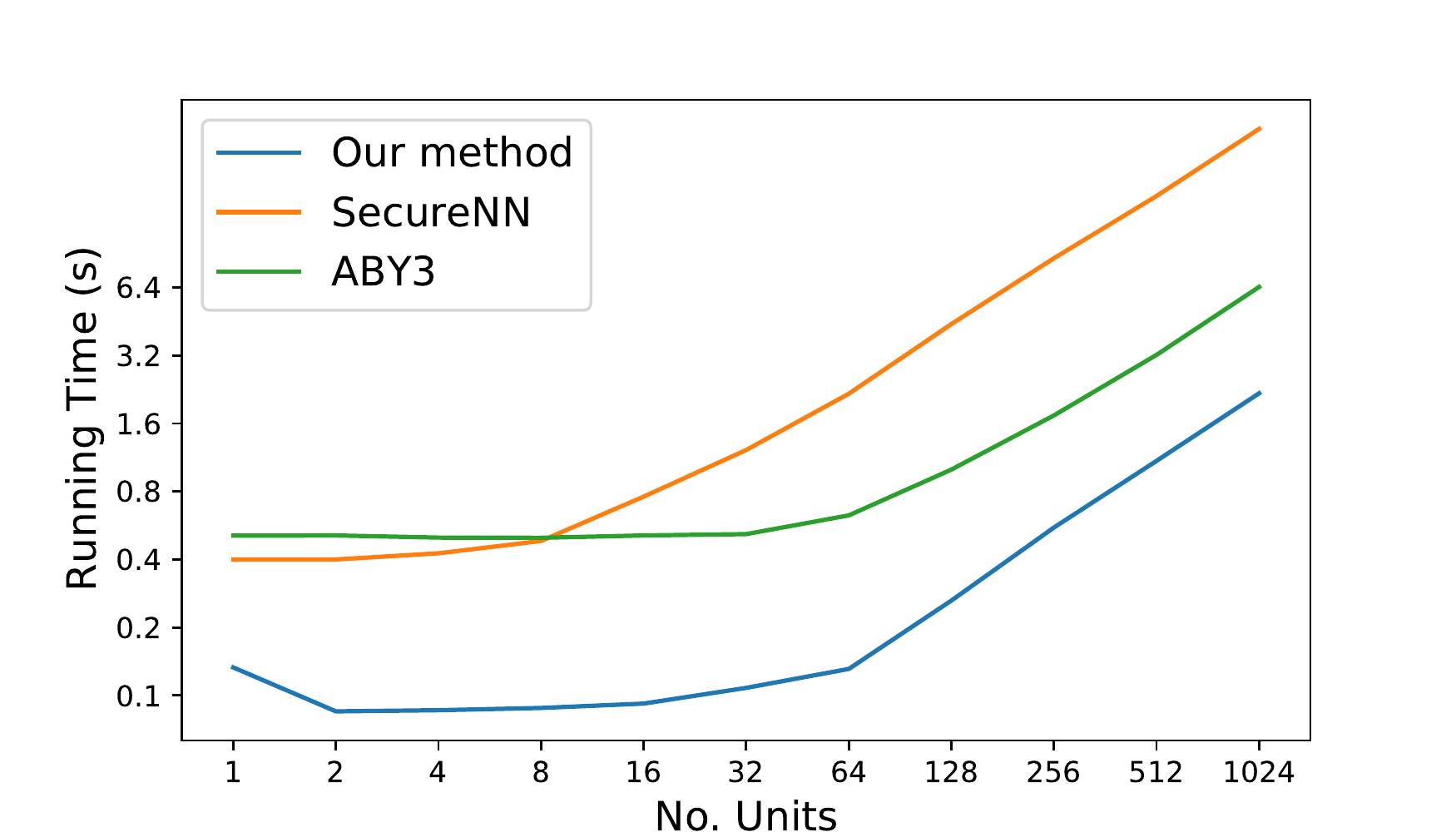}
    \caption{Running time}
    \end{subfigure}
    
    \begin{subfigure}[b]{0.4\textwidth}
    \centering
    \includegraphics[width=\textwidth]{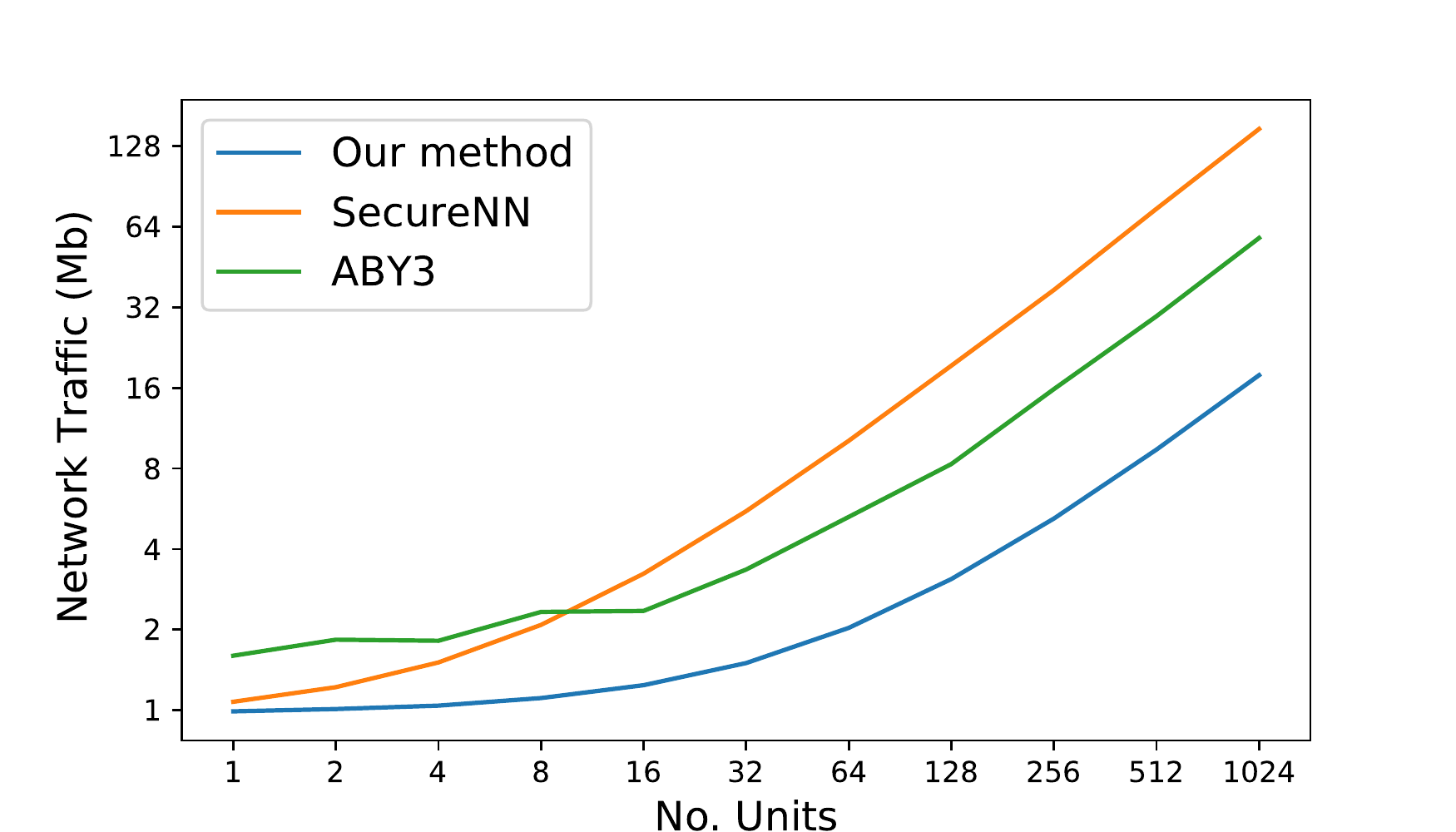}
    \caption{Total network traffic}
    \end{subfigure}
    
    \begin{subfigure}[b]{0.4\textwidth}
    \centering
    \includegraphics[width=\textwidth]{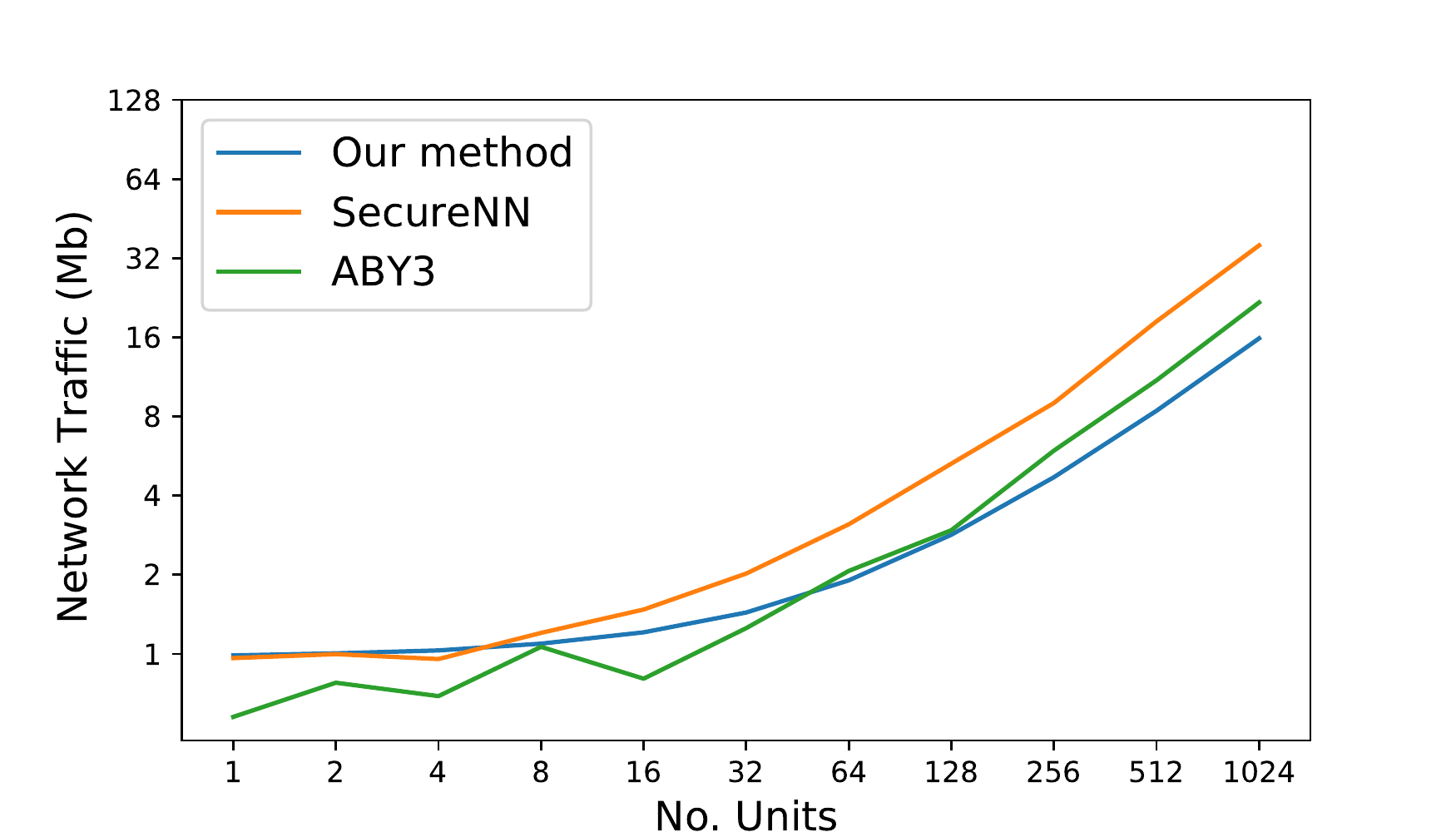}
    \caption{Network traffic between $P_0$ and $P_1$}
    \end{subfigure}
    
    \begin{subfigure}[b]{0.4\textwidth}
    \centering
    \includegraphics[width=\textwidth]{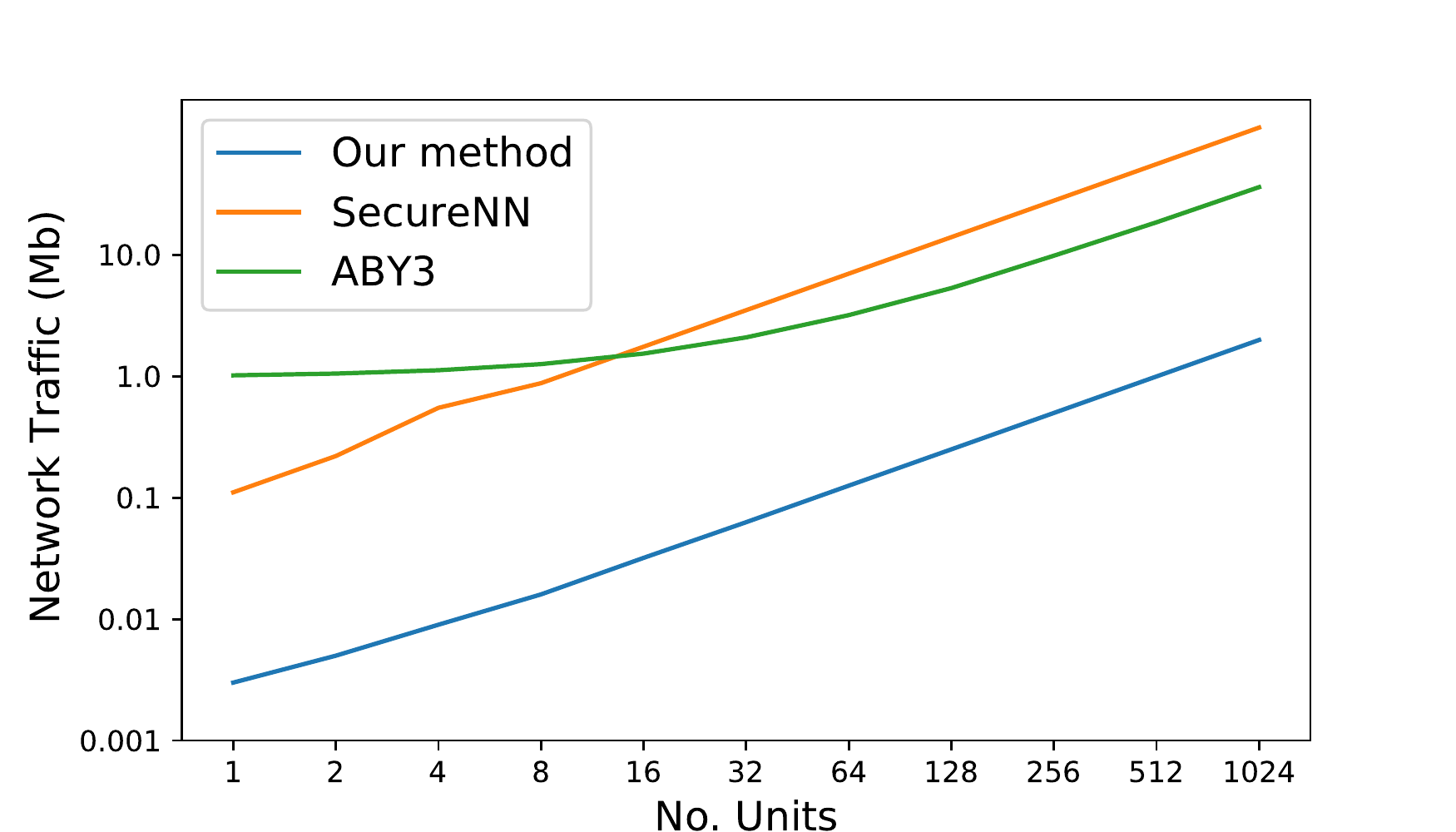}
    \caption{Network traffic of $P_2$}
    \end{subfigure}

\caption{Running time and network communication on fully-connected layers with different number of units. Blue: Our method; Orange: SecureNN; Green: ABY3.}
\label{fig:hidden-time-traffic}
\end{figure}

\nosection{Hidden Layers of Different Size}
In order to demonstrate the advantage of our method on non-linear activation functions,
we compare our method with ABY3 and SecureNN on fully connected layers of different numbers of hidden units (i.e., output dimension), and report the results in Figure \ref{fig:hidden-time-traffic}, where the input dimension is fixed to 1,000.
The result shows that our method tends to have higher speedup and more communication reductions with the increase of hidden units. 
The reduction of network traffic increases from 1× to 8× compared to SecureNN and 1.6× to 3.2× compared to ABY3 when the size of layer increases from 1 to 1024.
This is because that more units require more non-linear computations.
The portion of non-linear computations will be even larger in more complex models like convolutional neural networks. 
Hence, our method is potentially better on those models.

\subsection{Experiments on Real-World Datasets}
We also conduct experiments of logistic regression and neural networks on real-world datesets.
% Dataset
We train a logistic regression model on the The Gisette 
dataset\footnote{https://archive.ics.uci.edu/ml/datasets/Gisette}. 
Gisette dataset contains 50,000 training samples and 5,000 validation samples of dimension 5,000, with labels of -1 or 1. 

% Model Description
As for neural networks, we train two neural networks on the MNIST~\cite{lecun1998mnist} dataset. 
The MNIST dataset contains 50,000 training samples and 5,000 validation samples of dimension 28×28, with labels of 0 to 9, which we convert into one-hot vectors of dimension 10. 
We use ReLU for hidden layers and Sigmoid for output layers as activation functions.
The first neural network has a hidden layer of size 128, while the second neural network has two hidden layers of size 128 and 32.

\begin{figure}[t]
\centering
\includegraphics[width=6.5cm]{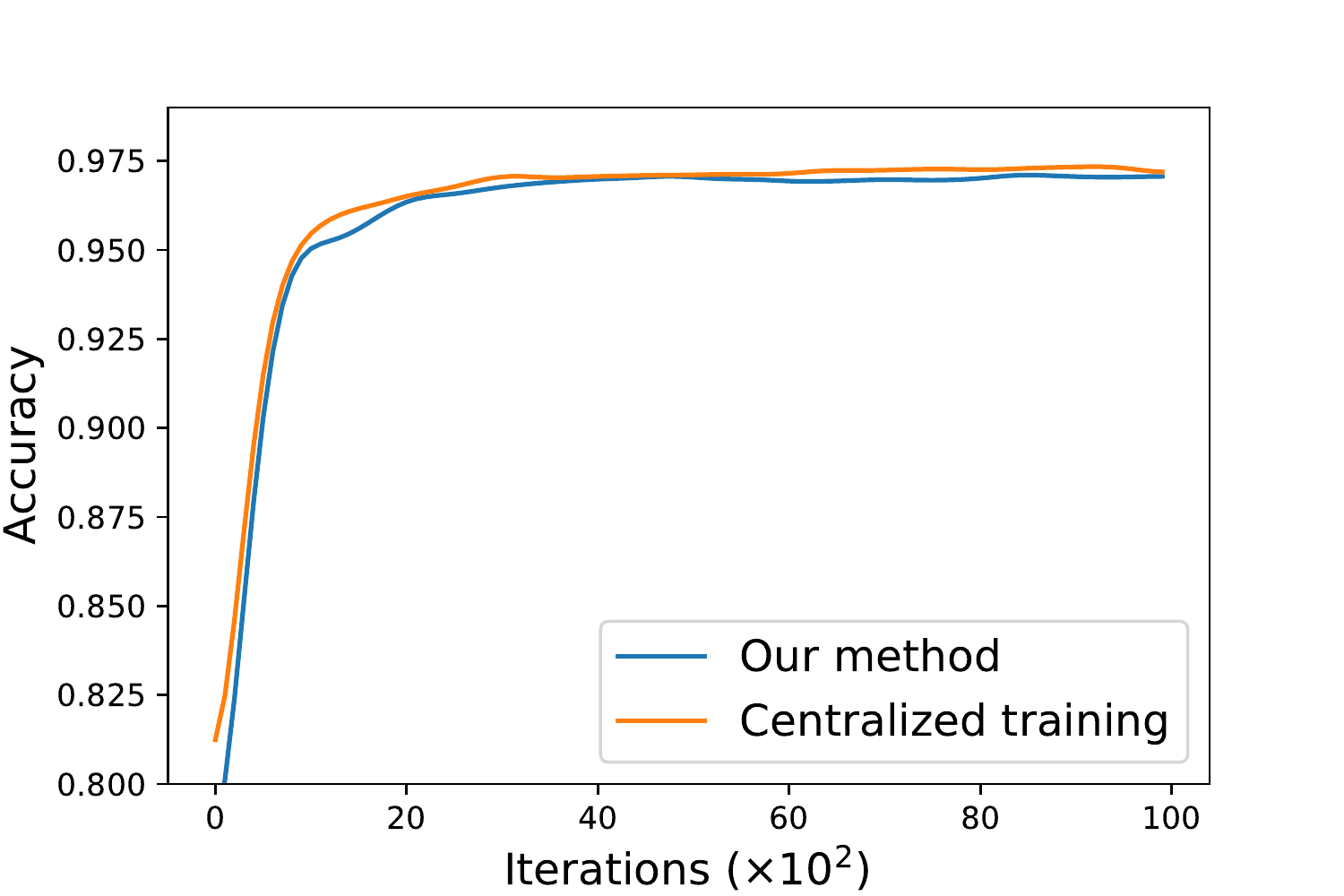}
\caption{Validation accuracy of logistic regression on Gisette dataset. Orange: Centralized plaintext model; Blue: Our method.}
\label{fig:lr-train}
\end{figure}

\begin{figure}[t]
\centering
\includegraphics[width=6.5cm]{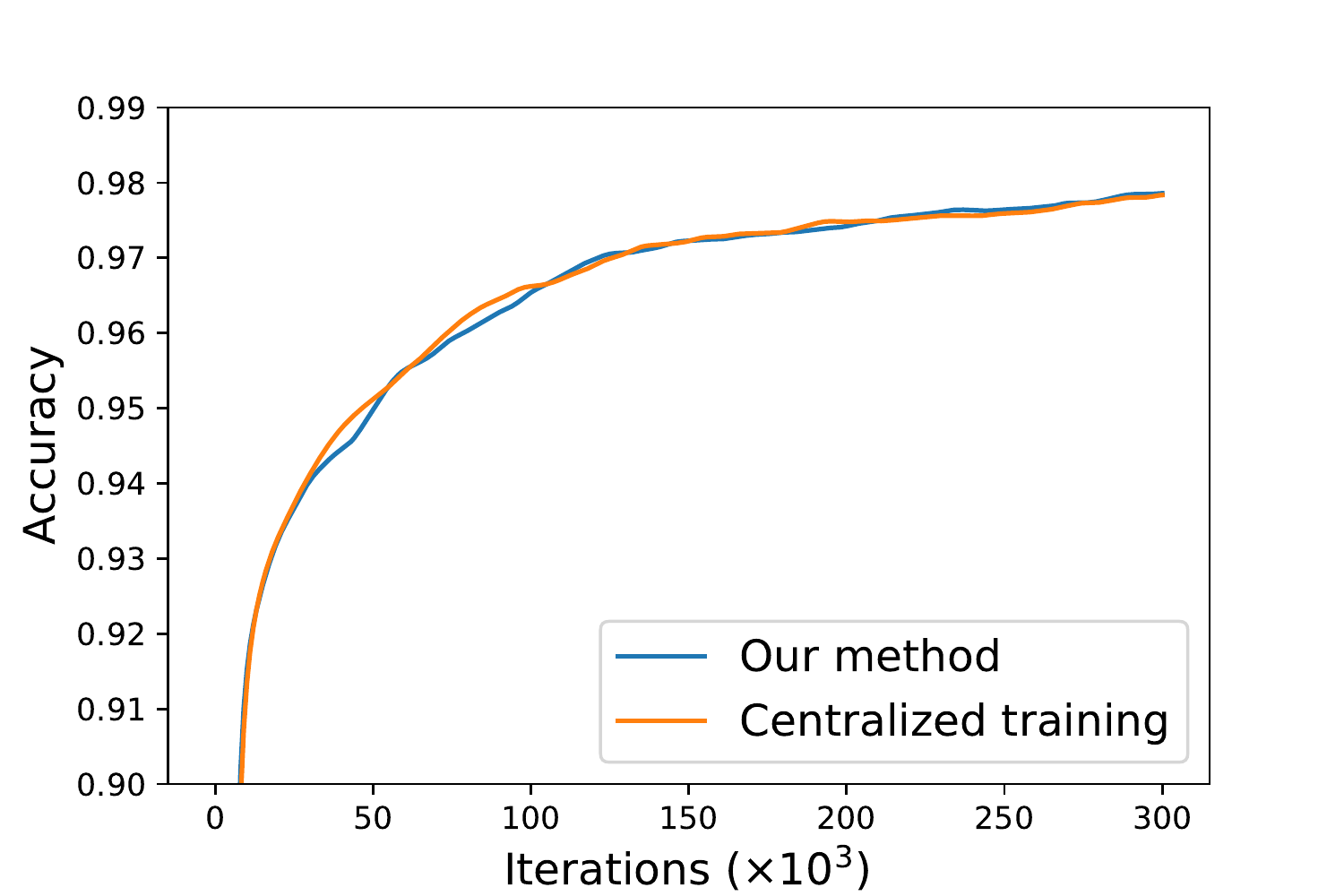}
\includegraphics[width=6.5cm]{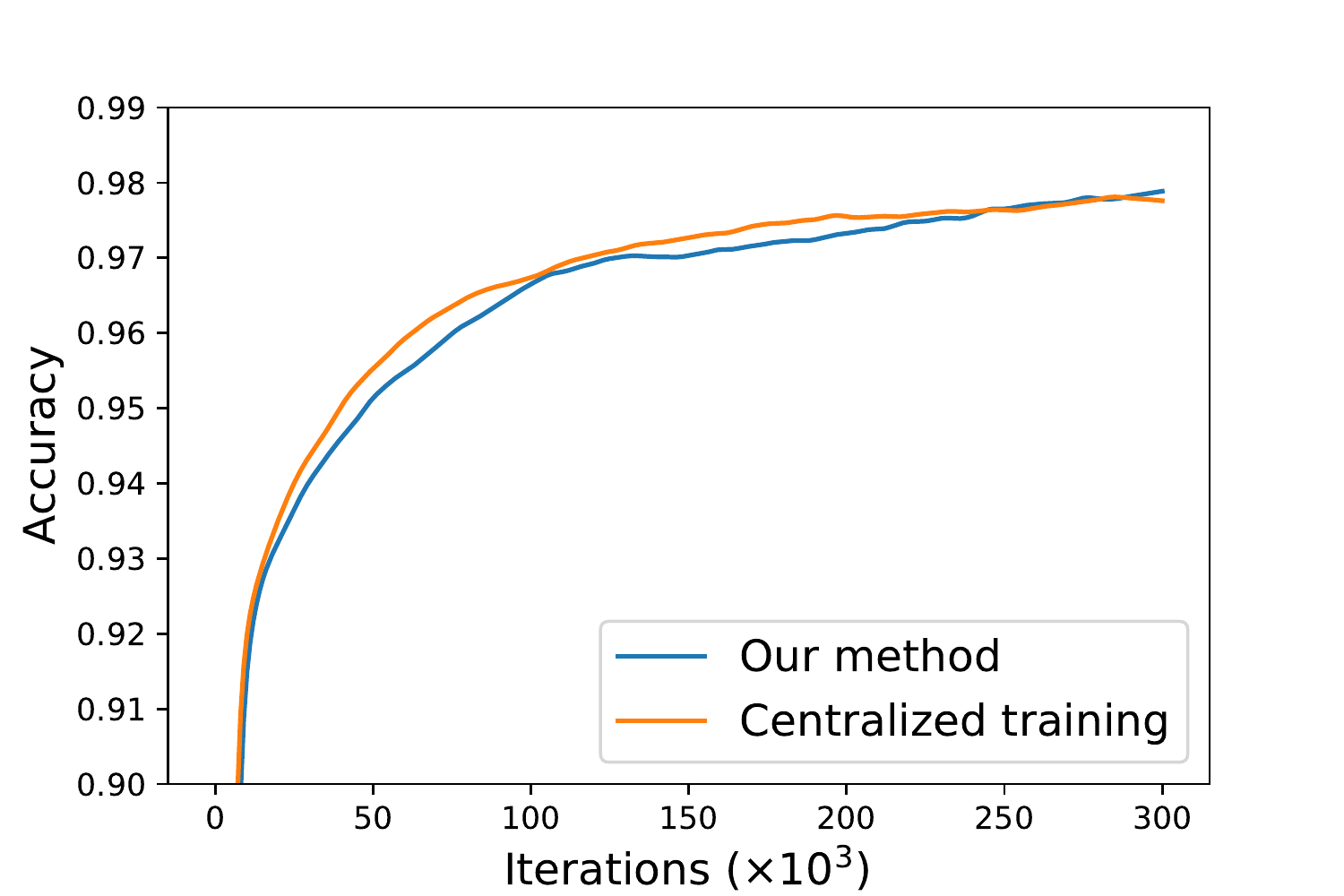}
\caption{Validation accuracy of neural network 784-128-10 and 784-128-32-10 on MNIST dataset. Orange: Centralized plaintext model; Blue: Our method.}
\label{fig:nn-train}
\end{figure}

\nosection{Model Performance}
We compare our method with the centralized plaintext training and report the accuracy curve of the logistic model in Figure \ref{fig:lr-train} and two neural networks in Figure~\ref{fig:nn-train}, respectively. 
The curves of our method and the centralized plaintext training are almost overlapped for all the three models, indicating that our method does not suffer from accuracy loss. 
This is because the main accuracy loss for our method is the conversion between float-point and fixed-point.  
However, since we use 64-bit fixed-point integer and precision bits of 23, this loss is very tiny and even negligible for machine learning models.

\begin{figure}[t]
\centering
\includegraphics[width=7cm]{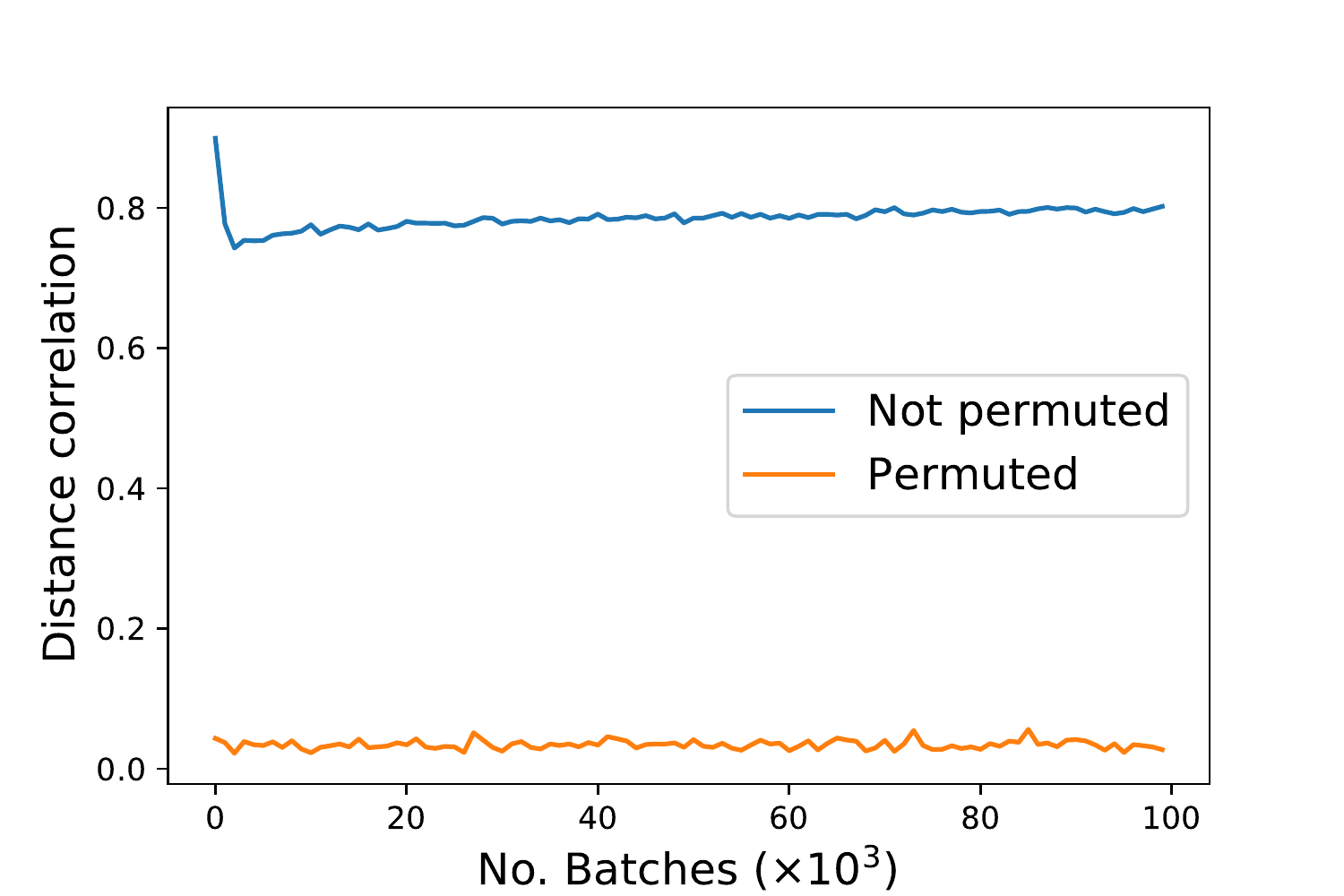}
\caption{Distance correlation between leaked data and original data. Blue: Not permuted; Orange: Permuted.}
\label{fig:nn-dcor}
\end{figure}

\nosection{Privacy Leakage}
We measure the distance correlation between original training data and the permuted hidden representation  (which is obtained by $P_2$) of the above neural network models, compare it with the no-permutation case (like split learning), and report the result in Figure \ref{fig:nn-dcor}.
The result shows that without permutation, the distance correlation is at a high value of about 0.8. 
After applying permutation, the distance correlation decreases to about 0.03, which indicates that almost no information about original data is leaked. 

\nosection{Simulated Attack}
We simulate the histogram attack proposed in Section \ref{section:security}.
The setting is that the adversary has 3,000 leaked images with the same distribution as the original MNIST training dataset and also has 3,000 hidden representations with dimension 128.
We extract ten hidden representations of digit 1 and find the most similar 10 samples via comparing the histogram distances with the leaked images.

\begin{figure}[t]
\centering
\includegraphics[width=6cm]{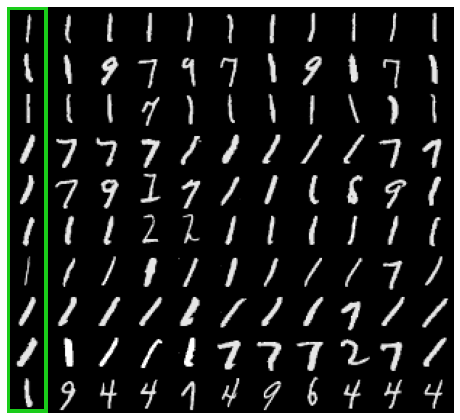}
\caption{The hidden representations are revealed without permutation. The left column is the original image, and the other 10 columns are the most similar samples find in the leaked data.}
\label{fig:attack1}
\end{figure}

Figure \ref{fig:attack1} is the result of the histogram attack when the adversary directly gets the hidden representations. 
The similar images to digit 1 found by the adversary are almost the same as the original image. Moreover, the thickness and the rotation angles of these similar images are close to the original image.

\begin{figure}[t]
\centering
\includegraphics[width=4cm]{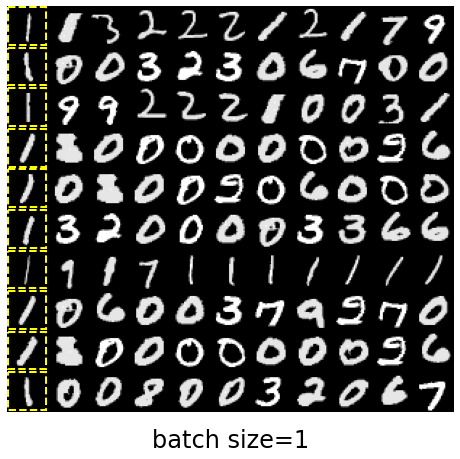}
\includegraphics[width=4cm]{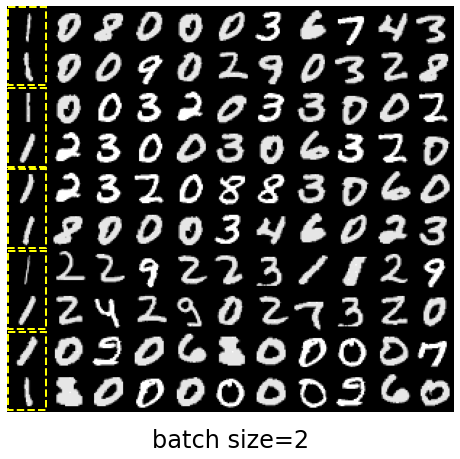}
\includegraphics[width=4cm]{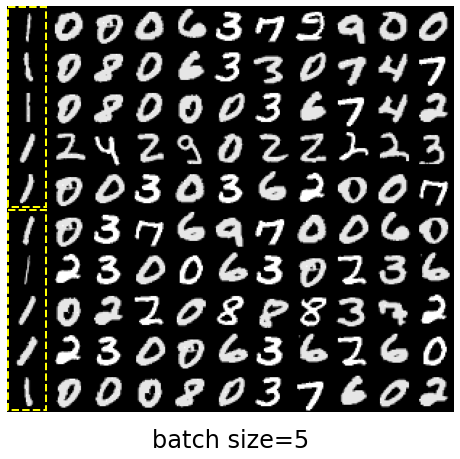}
\includegraphics[width=4cm]{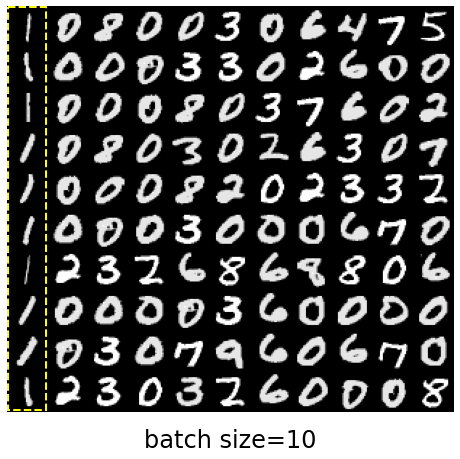}
\caption{Permutation is performed with batch size in \{1, 2, 5, 10\}. Each sample's hidden representation is shuffled within its batch (represented by yellow dashed rectangles).}
\label{fig:attack1-permuted}
\end{figure}

Figure \ref{fig:attack1-permuted} shows the result when permutation is applied in different batch sizes.
When batch size is one, it seems that the attack succeeds in the 7-th hidden representation. 
A possible reason is that the 7-th image is very dark since the white pixels are rare, causes the absolute values of elements in hidden representations very small.
When batch size is 1, the set of elements corresponding to the sample is unchanged after permutation and their absolute values are still very small. 
Hence, through histogram attack, images with large portion of black pixels are found, and most of them are of digit 1.
However, when the batch size is more than 1, multiple samples are shuffled together, the result tends to be completely random and has no relation with the original image.

\section{Conclusion and Future Work}
% Work done
In this paper, we propose a privacy-preserving machine learning system via combining arithmetic sharing and random permutation. 
% Main idea
% Algorithm
We exploit the element-wise property of many activation functions, and use random permutation to let one party do the computation without revealing information about the original data.
% Result
Through this, our method achieves better efficiency than state-of-the-art cryptographic solutions.
% Security
We adopt distance correlation to quantify the privacy leakage, illustrating that our method leaks very little information about the original data, while other non-provable secure methods leak information highly correlated to the original data.

In the future, we would like to further apply random permutation to other privacy-preserving machine learning models. We are also interested in developing practical secure methods under other security settings such as the malicious secure setting.

% With the demands for privacy-preserving machine learning continuously increase in recent years, 
% it is essential to achieve a balance between security and efficiency for practical applications. 
% %
% Our method adopts the semi-honest 3PC setting.
% But in practical applications, there are many limitations, e.g., two data sources do not trust any one else, or they require security against malicious adversaries.
% %
% Since cryptographic methods are still costly, it is worth researching to develop PPML systems that are practical secure under various (usually more strict) settings.
%

\section{Acknowledgement}
This work was supported in part by the National Key R\&D Program of China (No. 2018YFB1403001) and National Natural Science Foundation of China (No. 62172362, No. 72192823).

\bibliographystyle{elsarticle-harv}
\bibliography{main.bib}
\appendix
\end{document}